\pdfoutput=1

\documentclass[11pt]{article}

\usepackage[final]{acl}

\usepackage{times}
\usepackage{latexsym}
\usepackage[T1]{fontenc}
\usepackage[utf8]{inputenc}
\usepackage{microtype}
\usepackage{inconsolata}
\usepackage{xspace}
\usepackage{amsmath}
\usepackage{comment}
\usepackage{booktabs}
\usepackage{multirow}
\usepackage{enumitem}
\usepackage{color,soul}
\usepackage{tabularx}
\usepackage{graphicx}
\usepackage{algorithm}
\usepackage{algpseudocode}
\usepackage{adjustbox}
\usepackage{cleveref}
\usepackage{makecell}

\newcommand{\sparagraph}[1]{\vspace{0.0mm}\noindent\textbf{#1.}}

\newcommand{\QP}{\textsc{IP}\xspace}
\newcommand{\QPI}[1]{\textsc{IP}$_{#1}$}
\newcommand{\llm}{\textsc{GPT-4}\xspace}
\newcommand{\gpt}{\textsc{GPT-4}\xspace}
\newcommand{\partialctr}{\textsc{partial-ctr}\xspace}
\newcommand{\fullctr}{\textsc{full-ctr}\xspace}
\newcommand{\ee}{\textsc{Explore-Exploit}\xspace}
\newcommand{\randomdrop}{\textsc{random-ctr}\xspace}
\newcommand{\nodrop}{\textsc{no-drop}\xspace}
\newcommand{\example}[1]{{``\emph{#1}''}\normalsize}
\newcommand{\llmacc}{\textsc{LLM Acc.}\xspace}
\newcommand{\humanaggr}{\textsc{Agreement}\xspace}
\newcommand{\ctr}{\textsc{CTR}\xspace}
\newcommand{\ctravg}{\textsc{Avg. CTR}\xspace}

\title{Generative Explore-Exploit: Training-free Optimization of Generative Recommender Systems using LLM Optimizers}

\author{Lütfi Kerem Senel$^{1,2}$\thanks{~~Work done during an internship at Amazon.} ~~~~ Besnik Fetahu$^3$ ~~~~ Davis Yoshida$^3$ ~~~~  Zhiyu Chen$^3$ \\ 
 \textbf{Giuseppe Castellucci$^3$ ~~~~  Nikhita Vedula$^3$ ~~~~ Jason Choi$^3$ ~~~~ Shervin Malmasi$^3$} \\
 $^1$ Center for Information and Language Processing (CIS), LMU Munich, Germany \\
 $^2$ Munich Center for Machine Learning (MCML), Germany \\
 $^3$ Amazon.com, Inc. ~~~ Seattle, WA, USA \\
 \texttt{lksenel@gmail.com} \\
\texttt{\{besnikf,dayosh,zhiyuche,giusecas,veduln,chojson,malmasi\}@amazon.com}}

\begin{document}
\maketitle

\begin{abstract}
Recommender systems are widely used to suggest engaging content, and Large Language Models (LLMs) have given rise to generative recommenders.
Such systems can directly generate items, including for open-set tasks like question suggestion.
While the world knowledge of LLMs enable good recommendations, improving the generated content through user feedback is challenging as continuously fine-tuning LLMs is prohibitively expensive.
We present a training-free approach for optimizing generative recommenders by connecting user feedback loops to LLM-based optimizers.
We propose a \textit{generative explore-exploit} method that can not only exploit generated items with known high engagement, but also actively explore and discover hidden population preferences to improve recommendation quality. 
We evaluate our approach on question generation in two domains (e-commerce and general knowledge), and model user feedback with Click Through Rate (CTR).
Experiments show our LLM-based explore-exploit approach can iteratively improve recommendations, and consistently increase CTR.
Ablation analysis shows that generative exploration is key to learning user preferences, avoiding the pitfalls of greedy exploit-only approaches.
A human evaluation strongly supports our quantitative findings.
\end{abstract}

\section{Introduction}\label{sec:introduction}
\begin{figure}[ht!]
    \centering
    \includegraphics[width=\columnwidth]{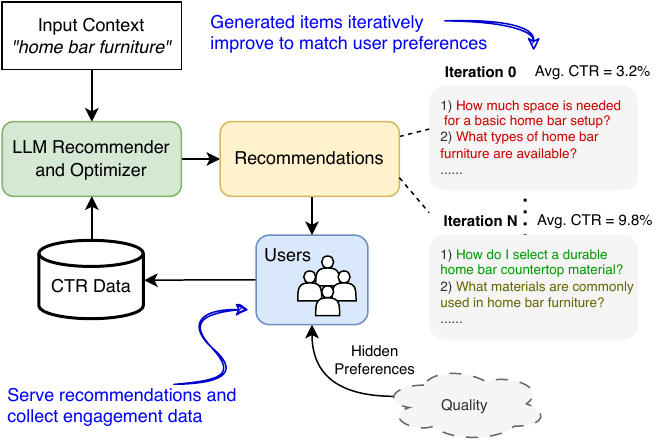}
    \caption{Overview of our generative recommender approach. It iteratively refines its item pool using feedback signals based on clicks to gradually improve the relevance of the questions to its user base.}
    \label{fig:example_questions}
\end{figure}
Recommender systems are widely used for various applications, including suggesting items in catalogs (music, books, videos)~\cite{DBLP:journals/corr/abs-2310-10108}, e-commerce~\cite{10.1145/3604915.3608836}, and search results~\cite{DBLP:conf/sigir/Najork23,metzler2021rethinking}. A core part of a recommender system is improving the relevance of recommended items based on the user's preferences.
Such preferences can be either \textit{explicit} (e.g. user provided preferences about a brand in e-commerce queries), or \textit{implicit} based on engagement or Click Through Rate (CTR) of provided recommendations. Furthermore, the more users interact with a recommender system, the more the recommended items are optimized to match such explicit or implicit user preferences.

Traditional recommender systems use optimization approaches such as (deep) collaborative filtering~\cite{DBLP:journals/access/MolaeiH0J21}, matrix factorization~\cite{sarwar2002incremental}, and reinforcement learning~\cite{DBLP:journals/csur/AfsarCF23}.
A common thread is that they are all applied to a static set of items.
However, the increasing capabilities of Large Language Models (LLMs) has led to the development of \emph{generative recommender systems} \cite{zheng2023generative, li2023gpt4rec}.
Generative recommenders can be used to directly generate items or text content for recommendation. 
The immense world knowledge of LLMs makes them excellent at certain generative recommendation tasks such as question or query suggestion \cite{vedula2024question}.
Unlike traditional recommender systems, they do not choose from a fixed item set, and the range of valid items that could be generated is vast.

We investigate how LLM-based generative recommenders can be iteratively optimized based on implicit user feedback, e.g. by following CTR signals (cf. \Cref{fig:example_questions}).
Such recommenders needs to support very large numbers of input contexts (e.g. items or queries).
However, fine-tuning them to improve recommendations is prohibitively costly given their size.
Our approach is training free: improving item recommendations does not need fine-tuning; instead we propose novel methods to synthesize context-specific implicit engagement signals (e.g. CTR) as part of the LLM input. We also propose a \textit{generative explore-exploit} mechanism to generate and evaluate new candidate items.
Our work is applicable to numerous applications such as search~\cite{DBLP:conf/sigir/Najork23,metzler2021rethinking}, question answering~\cite{huber-etal-2022-ccqa,qin-etal-2023-webcpm}, and query/question suggestion features in information seeking systems \cite{DBLP:conf/kdd/MitraJVG21}.

To demonstrate our approach, we focus on the specific task of Question Generation (QG) \cite{10.1145/3372923.3404786}.
This involves suggesting engaging (e.g. high CTR) questions to help users explore a topic.
Experiments show that by leveraging both generative exploration and exploitation, our approach can adapt its recommendations to match the preferences of a population whose preferences are not directly observable.
We show that a feedback loop based on CTR can successfully guide the LLM to \emph{explore} new questions types and topics, and \emph{exploit} previously generated ones. Finally, we show that our approach converges \emph{faster} to questions that meet user needs, compared to baselines without CTR signals.

Our work makes the following contributions:
\begin{itemize}
    \item To the best of our knowledge, we are the first to propose a training-free \emph{generative recommender} approach, which optimizes its recommendations using a feedback loop based on implicit user feedback such as CTR.
    \item We create an offline experimental framework to simulate user preferences and their click behaviors to enable efficient development.
    \item A detailed evaluation on e-commerce and general knowledge domains demonstrates the effectiveness of our approach.
    We show that generative exploration and the use of prior CTR performance data are key elements of improving LLM-generated recommendations.
\end{itemize}

\section{Related Work}
\label{sec:related_work}

\sparagraph{LLMs for Recommendation} 
Recent research has been exploring generative recommender systems \cite{zheng2023generative, wang2023generative, li2023gpt4rec, li2023large}, by integrating LLMs into different stages~\cite{lin2023can} such as feature augmentation \cite{xi2023towards, wu2023leveraging}, representation enhancement \cite{li2023prompt,rajput2023recommender}, item scoring \cite{zhang2023collm,tang2023one}, and user interaction \cite{dong2023musechat,yang2021improving}.
Most of these studies follow a retrieval-based paradigm, either scoring and recommending existing candidate items or generating a ranked item list grounded to an existing item pool. One shortcoming of such works is that they may fail to satisfy the needs of diverse or unseen users. Furthermore, they do not easily adapt to evolving domains or applications such as growing product catalogs or modified article contents. A solution to these issues is to generate the items for recommendation~\cite{zheng2023generative, wang2023generative}. We follow a similar approach, but differ in important aspects such as we do not require any training to generate content (i.e. questions) that are relevant to latent user needs.

Prior work has employed LLMs within recommender systems for CTR prediction, aiming to estimate the probability of users engaging with items \cite{fu2023unified,li2023ctrl}. Similar to us, \citet{liu2023chatgpt} designed specific zero-shot and few-shot prompts to assess how LLMs predict recommendation ratings.
To our knowledge, we are first to use LLMs to simultaneously generate recommendations and optimize CTR in the same framework. 
We do this by building upon the findings of \citet{yang2023LLMsOptimizers}, 
by leveraging LLMs ability to understand and make use of the connection between textual input (questions in this work) and its performance on a target task. Our approach learns to generate questions that can achieve higher CTR.

\sparagraph{Question Generation}
QG is important in various NLP applications \cite{mulla2023automatic} such as reading comprehension \cite{ghanem-etal-2022-question}, conversational recommendation \cite{kostric2021soliciting}.
We specifically focus question suggestions, i.e., questions that \textit{users might want to ask} a system.
A notable use case is People Also Ask (PAA) questions in web search \cite{10.1145/3372923.3404786}, which suggests users potential next questions they can ask to further explore a topic.
Unlike PAA where questions are typically static, we generate novel and more engaging suggestions based on engagement signals from users interactions (clicks). This helps create questions aligned with the interests of the user population.

\sparagraph{User Click Simulation} Several recommender systems studies have developed ranking models to simulate user click behavior in the web search domain \cite{dupret2008UBM, zhang2023reranking}. %
Recently, \citet{wang-etal-2023-rethinking-evaluation} prompted LLMs with personas and rules to simulate user behavior within conversational recommender systems.
Inspired by their work, we design a user simulator that is representative of real-world user cohorts with varying personas and needs, and use LLMs to evaluate the question relevance to these personas.
\section{Problem Definition}
\label{sec:problem_definition}

Given a population of users $U=\{u_1,\ldots, u_n\}$, a user $u$ may search for multiple topics $t_1,\ldots,t_k$ (e.g. ``\emph{Biology}'', ``\emph{Smartphones}'').
The task is to generate a fixed set of questions to be suggested for a topic $t$, referred as \textit{item pool} (\QP), such that the questions are likely to be of interest to as many users as possible in $U$ that have searched for $t$.
Actual user interests about $t$ are \emph{hidden} and can only be observed through interactions (e.g. clicks) with the generated items.

Using a sufficiently large LLM with reasonable instruction following capabilities, we want to generate \QP for $t$, which is a list of $N$ questions that maximize Click Through Rate (CTR) from all users  $u \in U$ that have interests in $t$.

\section{Generative Recommender Approach}
\label{sec:approach}

\begin{figure*}[thb!]
    \centering
    \includegraphics[width=1.\textwidth]{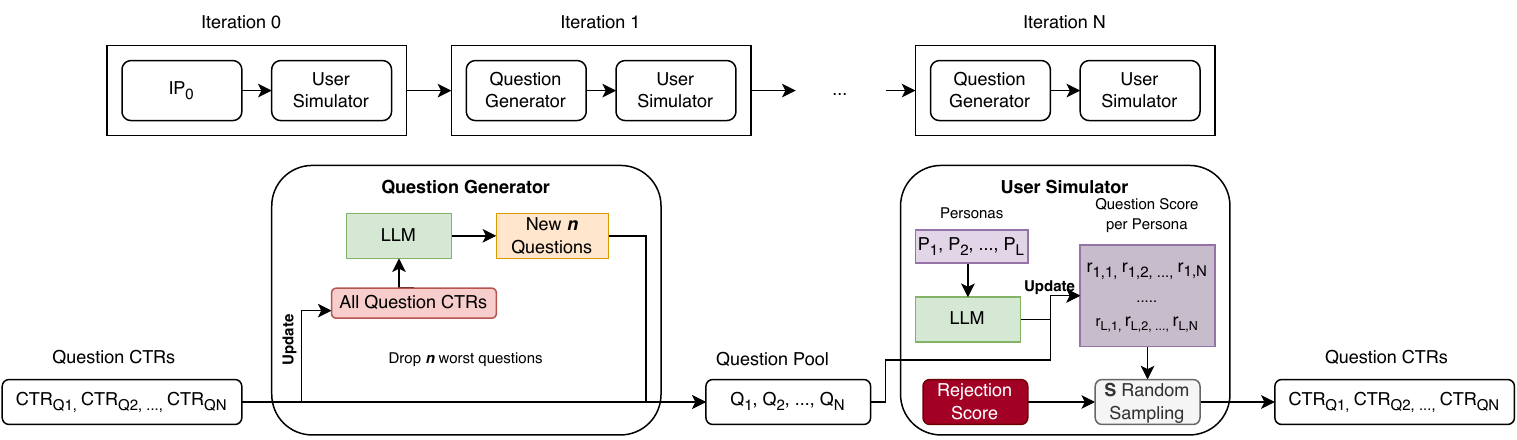}
    \caption{Overview of our training-free generative recommendation approach. Our approach generates a question pool that has maximal relevance to its underlying user population base. Without any explicit signal on what the user's interests are, it exploits click through rate (\ctr) of questions to iteratively refine what question shapes and about what aspects are generated. Initially, in the first iteration the questions are unlikely to be relevant to its user base, however, as \ctr signal is gathered across multiple rounds of feedback iterations, our approach is able to progressively improve the question relevance.}
    \label{fig:overview}
\end{figure*}

Our training-free generative recommender approach is outlined in \Cref{fig:overview}.
At a high level, our proposed approach iteratively improves LLM-generated outputs as follows.
An LLM is used to generate the initial candidate items for a given task, and the click-through rates (CTR) of these items are then measured to gauge user engagement and preference. Based on these observed CTRs, a prompt is constructed to encapsulate this performance data. This prompt serves as a basis for applying a dual approach of generative exploration and exploitation using an LLM optimizer via in-context learning. The optimizer refines the generated items by balancing the exploration of new possibilities and the exploitation of known high-performing elements, to enhance the quality and engagement of the generated content.
Our method is suitable for tasks where many valid generations are possible, like summarization and question generation.

Our approach is iterative, with user interaction data collected in each round $i$ in order to refine the \QPI{i} on a given topic $t$.
More specifically, in each iteration: (i) the $n$ worst items are dropped from \QP, (ii) $n$ new items are generated and added to \QP, and (iii) the performance of items in the updated \QP is observed via an interaction feedback loop.
Items in the $i$-th iteration are indicated by \QPI{{i}}.

\sparagraph{Item Pool (\QP) Initialization}
In the first iteration, only the target topic of interest $t$ is known.
Hence, we initialize the item pool \QPI{0}, by simply generating relevant questions for $t$ (prompts are shown \Cref{fig:initial_prompts}).
These questions are recommended to users, and their interactions (clicks) are used to refine \QPI{i} in subsequent iterations to improve CTR.

\sparagraph{Iterative Refinement using CTR} We iteratively refine \QP based on the CTR signal; in each iteration, \QP is updated by dropping low CTR items and adding new questions that are optimized to increase the overall CTR of \QP. To improve the CTR we rely on the ability of LLMs to act as an optimizer for a target task by following instructions, without requiring any fine-tuning \cite{yang2023LLMsOptimizers}. We iteratively \emph{update} the input instructions to the LLM by including previously generated items and their observed population CTRs. This allows LLMs to optimize their output based on the instructions, with the CTR values providing the LLM with both positive and negative instances of good as well as bad (low \ctr) question.
With sufficient iterations, the LLMs are able to converge to an \QP that maximizes CTR.
To do this, we propose two approaches: (i) \textbf{\fullctr} and (ii) \textbf{\ee}.

\paragraph{\fullctr:} Here, we provide the LLM all previously generated questions\footnote{We experimented with providing only the latest state of \QP, however, this caused LLM to re-generate previously generated and dropped questions.} along with their observed CTR scores
As part of the prompt, the LLM is instructed to optimize its generated questions such that the questions in \QPI{i+1} will obtain high CTR. Similar to \citet{yang2023LLMsOptimizers},
\Cref{fig:generation_prompts} provides the prompt used to iteratively refine \QP.

\paragraph{\ee:}
The main shortcoming of \fullctr
is that it can only exploit observed user preferences, and biases generation towards these.
This greedy approach can prevent the optimizer from exploring different generations that may fail, thus limiting overall quality.
We overcome this with our \ee strategy, where in every iteration we drop the worst questions, and generate \emph{two} sets of questions as follows.
First, in an \textit{explore} phase, a set of $n$ questions is generated by providing only \QPI{i}, without any CTR values.
Second, similar to \fullctr, a set of $n$ questions is generated using an \emph{exploit} prompt, designed to generate questions that are on the same topic as the best performing question in \QPI{i} (See Figure~\ref{fig:generation_prompts} for the prompt).
To avoid saturating \QPI{i+1} with only questions on one topic, \ee instructs the LLM to explore new topics for which to generate new questions that increase the diversity of the entire \QP set.

\section{User Click Simulator}
\label{sec:user_simulator}

Developing and evaluating our approach requires user engagement data, which is expensive to collect, and involves complex privacy concerns. To enable fast offline experimentation, we simulate  feedback by modeling user preferences and their click behavior on suggested questions using LLMs.

We represent users with different interests and goals via specific prompts (see \Cref{tab:Personas} for details).
These user personas $p$ have specific interests for different topics.
They do not represent a single user, but rather a population of users with similar characteristics.
We split the simulator into two steps: (i) relevance scoring, and (ii) action simulation.

\subsection{Relevance Scoring}\label{subsec:relevance_scoring}

For a question $q_i$ generated for a topic $t$, we evaluate its relevance to a persona $p_j$ by prompting an LLM to score $q_i$ on a scale $r\in \{1,\ldots, 10\}$:
\begin{equation*}
   r_{i,j} = \textsc{qs}(q_i, p_j, t)
\end{equation*}
The prompt used in \textsc{qs} is given in Figure~\ref{fig:scoring_prompts}. When using the LLM to compute $r$, during decoding we set the temperature to 1 to induce variation in the behavior of a persona, thus, mimicking how different users falling under the same persona would behave.
The score $r_{i,j}$ is computed once and independently from other questions in \QP. Independent scoring reduces any potential bias stemming from other questions in \QP.

\subsection{Action Simulation}\label{subsec:action_simulation}
\begin{figure}
    \centering
    \includegraphics[width=\columnwidth]{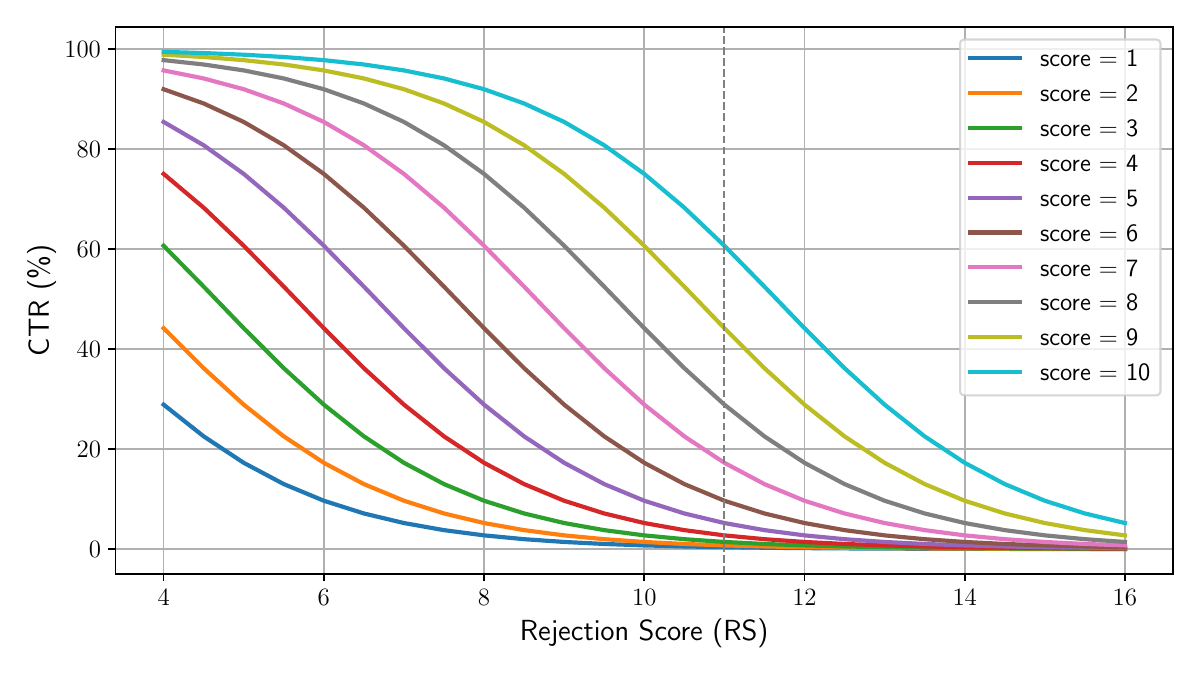}
    \caption{Theoretical CTR values with $T=1.5$ for varying $RS$ and for 3 shown questions ($K=3$) with equal scores ranging from 1 to 10. The dashed vertical line ($RS=11$) shows the rejection score used in our experiments.}
    \label{fig:RS}
\end{figure}

We obtain CTR values by simulating $S$ user interactions based on the pre-computed $r_{ij}$ values.
In each interaction, we uniformly sample a persona ($p_j$) and uniformly sample a set of $K$ questions from \QP. $p_j$ then takes one of $K+1$ actions: (i) clicking one of the $K$ questions, or (ii) not clicking anything.
For $q_i\in K$ and $p_j$ we model the probability of $q_i$ being clicked as a temperature $T$ softmax:
\begin{equation}\label{eq:softmax}
P\left(\textsc{click}| p_j, q_i\right) = \frac{
    \mathrm{e}^{r_{ij}/T}
}{
    \mathrm{e}^{RS/T} + \sum\limits_{q_k \in K} \mathrm{e}^{r_{kj}/T}
}
\end{equation}
where $RS$ represents a fixed ``\emph{rejection score}'', which is the logit for clicking on none of the options. Figure~\ref{fig:RS} shows the resulting CTRs for varying relevance scores and values of $RS$. Low $RS$ values (e.g. $RS < 10$) allows for unrealistic high CTR (e.g., +90\%) and high $RS$ values (e.g. $RS > 12$) suppress the CTR heavily, not leaving much room for improvement. We set  $RS = 11$ in our experiments, which yields potentially realistic CTR values while allowing for room to improve by increasing the relevance of suggested questions.

\section{Experimental Setup}\label{sec:setup}

Here, we discuss the experimental setup, namely the domains on which we assess our proposed approach and competitors. Furthermore, we explain in detail the personas used for experimentation, which aim at mimicking real user cohorts. Finally, we will define the evaluation metrics used.

\begin{table}[ht!]
    \centering
    \resizebox{1.0\columnwidth}{!}{
    \begin{tabular}{l|l|l}
    \toprule
    \textbf{Domain} & \textsc{E-Commerce} & \textsc{General Knowledge} \\
    \midrule
    \textbf{Source} & Product Category & Wikipedia Article \\
    \midrule
    \multirow{5}{*}{\makecell{\textbf{Sample}\\\textbf{Topics}}} & Spray Bottles & Stoicism \\
    & Home Bar Furniture & Tabata training \\
    & Cookware Sets & Friedrich Nietzsche \\
    & Lighting \& Ceiling Fans & Chernobyl disaster \\
    & TV Antennas & Artificial superintelligence \\
    \midrule
    \multirow{5}{*}{\textbf{Personas}} & Price & Discussion-Focused \\
    & Quality & History-Focused \\
    & Brand Reputation & Event-Focused \\
    & Features \& Functionality & Person-Focused \\
    & Ethical Considerations & Location-Focused \\
    \bottomrule
    \end{tabular}}
    \caption{For the two domains, we consider 50 product categories and Wikipedia articles. We list sample topics (see \Cref{sec:input_topics} for the complete list) and the corresponding user personas, whose interests are hidden from our approach.}
    \label{tab:domains}
\end{table}

\subsection{Data and Domains}
\Cref{tab:domains} provides an overview of the details for the two domains we experiment with. The table lists sample topics of interest, alongside the personas we experiment with. The complete list of topics for both domains is shown in \Cref{sec:input_topics}.

\noindent\textbf{E-Commerce:} Suggested questions in online shopping have high utility as they allow users to explore the product space and make informed purchase decisions.
The inputs are 50 random \emph{product categories} from the Amazon Review Dataset \cite{ni2019trec}, while the personas represent shoppers.

\noindent\textbf{General Knowledge:} The inputs are Wikipedia article titles, while the personas represent users interested in a particular aspect of the article. We randomly sample 50 featured Wikipedia articles ranging across different categories (see \Cref{sec:datasets_appx} for details).

\subsection{User Personas}

\noindent\textbf{E-Commerce Personas:} The personas are defined in terms of \emph{shopping preferences}, a common way to describe customer behavior~\cite{Carmel2020,10.1145/3604915.3608836}.
Inline with preferences proposed in literature, we use the following: \textit{Price}, \textit{Quality}, and \textit{Brand Reputation}, \textit{Features and Functionality}, and \textit{Ethical Considerations}.
We experiment with user populations that consist of only one persona type as well as populations that contain multiple personas.
To extract CTR for populations that include multiple personas, we first compute question relevance, $r$ (cf \S\ref{subsec:relevance_scoring}) for a persona and question pair, then consequentially for each user click simulation, we randomly select one of the personas.

\noindent\textbf{General Knowledge Personas:}
Defining personas for this domain is more challenging. This is mainly due to Wikipedia being very diverse, and user preferences can vary greatly depending on the Wikipedia article or category. 

Since personas,  theoretically can be generated independently per article and category, hence, being highly sparse and potentially representing an unrealistic scenario to experiment with, we simplify the process and consider personas based on their focus as shown in Table~\ref{tab:domains}
Such personas are general enough to be applicable to most Wikipedia articles, allowing us to gain representative insights.

\subsection{Approach Setup}\label{subsec:approach_setup}

We set the size of \QP to 5 questions, and the number of generated (and dropped) questions at every iteration to $n=1$.
We allow our approach to refine \QP for 15 iterations ($I=15$). 
The number of click simulations in \emph{each} iteration is set to $S=5000$; a persona is shown $K=3$ questions at each simulation.
We empirically set the \emph{softmax temperature} to $T=1.5$ and the rejection score to $RS=11$ (cf. \S\ref{sec:user_simulator}).
Additional details about the simulator setup are provided in \Cref{sec:simulator_details}.

\paragraph{LLM:}
For all experiments we use \gpt~\cite{openai2023gpt4}, considered the most capable LLM (at the time of writing).\footnote{The \texttt{gpt-4-1106-preview} model was used.}
For simplicity, the same LLM is used for question generation and the user simulator. All prompts are provided in \Cref{sec:Prompts}.

\subsection{Approach Configurations}
\label{subsec:baselines}

To evaluate the effectiveness of our proposed approach and its components, we compare against ablations that assess the impact of each component.

\noindent\textbf{\randomdrop:} Instead of using the CTR signal, it uses random CTR values (between $[0\%-15\%]$) for dropping the worst question and writing a new one. This ablation tests the impact of CTR signals.

\noindent\textbf{\nodrop:} expands the \QP up to $N$ iterations for which it is executed, without dropping any questions. The CTR value of the \nodrop at iteration $i$ reflects the performance of directly generating 20 questions (initially |\QP|=5, and maximum number of iterations is 15). This ablation highlights the usefulness of the iterative nature of our approach, where questions with lowest CTR are dropped.

\noindent\textbf{\partialctr:} uses CTR signal to drop the worst performing questions from \QP at every iteration, however, in the input instruction prompt we do not provide the \ctr that the previously generated questions obtained. In this way we can test the impact of \ctr signal in the instruction prompt and its outcome in terms of obtained \ctr by \QP.

\subsection{Evaluation Metrics}
\label{subsec:metrics}

We define metrics for two evaluations: (i) item relevance scoring, (ii) recommendation performance.

\noindent\textbf{Item Relevance Scoring:} To measure \humanaggr, we compute the agreement between human annotators when judging which question in a pair is more relevant for a persona.
We also compute \llmacc, which measures the \emph{accuracy} or \emph{alignment} of LLMs with the human judgment.\footnote{Here we consider only cases where the human annotators agree on the more relevant question for a persona.} Namely, we measure if the LLM assigns a higher score to the question from the pair that was judged by annotators as being more relevant.

\noindent\textbf{Recommendation Performance:} To assess the overall performance we compute the following metrics: (i) \ctr values across iterations, (ii) average CTR score across $N$ iterations, and (iii) human annotation, by comparing questions in \QPI{0} vs. \QPI{I}.

\section{Results}
\label{sec:evaluation}

We present the experimental results for the two components of our approach: (i) item relevance scoring, and (ii) recommendation performance.

\subsection{Relevance Scoring Results}
\label{subsec:question_scoring_eval}

\begin{table}[htb!]
    \centering
    \resizebox{\columnwidth}{!}{%
    \begin{tabular}{l l l l}
        \toprule
                                   &                & \humanaggr \%  & \llmacc \%    \\
       \midrule
       \multirow{4}{*}{$\Delta$ Score } & All            & 70.2\% (132/188) & 77.3\% (102/132) \\
                                   & 2              & 69.2\% (63/91)   & 71.4\% (45/63)   \\
                                   & 3              & 71.4\% (35/49)   & 91.4\% (32/35)   \\
                                   & 4              & 70.8\% (34/48)   & 73.5\% (25/34)   \\
       \midrule
       \multirow{3}{*}{\textsc{Preference}} & Ethical Cons.  & 73.2\% (30/41)   & 93.3\% (28/30)   \\
                                   & Feat. \& Func. & 71.2\% (57/80)   & 73.7\% (42/57)   \\
                                   & Quality        & 67.2\% (45/67)   & 71.1\% (32/45)   \\
       \midrule
       \multirow{3}{*}{\textsc{Category}}   & Cookware Sets  & 73.6\% (53/72)   & 90.6\% (48/53)   \\
                                   & Spray Bottles  & 68.3\% (43/63)   & 76.7\% (33/43)   \\
                                   & TV Antennas    & 67.9\% (36/53)   & 58.3\% (21/36)   \\
       \bottomrule

    \end{tabular}
    }
    \caption{\small Question scoring \humanaggr and \llmacc (cf. \S\ref{subsec:metrics}) on the E-Commerce domain. Results are broken down across three main categories: 1) based on the gap between the relevance scores of  questions pairs considered ($\Delta$ Score); 2) persona; and 3) product category. Number of agreed and correct pairs/number of pairs is shown in parentheses.
    }
    \label{tab:human_eval}
\end{table}

\Cref{tab:human_eval} shows the human evaluation results for the question relevance scoring (cf. \S\ref{subsec:relevance_scoring}). This assesses if LLMs can reliably be used to judge question relevance.
Due to the more complex nature of the e-commerce domain, we only evaluate relevance scoring on this domain.
We sample three different personas from three different product categories, and randomly pair questions across different iterations. This results in 188 pairs for evaluation. Each pair is judged by two expert annotators.\footnote{We recruited expert internal annotators who were trained on the provided evaluation protocol on how to determine when a question is better and more suited for a persona.}

In 70.2\% of cases human annotators agree on which question for a given pair is more relevant. 
In the subset of question pairs with human agreement, the \llmacc score is 77.3\%.
This means that LLMs correctly assign higher relevance scores to questions that are also judged by annotators as being more relevant for a persona. Note here that we only show \llmacc for the cases where there is human agreement, since we assume that there is no ambiguity and thus can objectively assess LLM's alignment with human annotators.
Across different product categories, \humanaggr relatively consistent and varies between 67.9\% and 73.6\%, while \llmacc changes more drastically between 58.3\% (\textit{TV Antennas}) and 90.6\% (\textit{Cookware Sets}).

In sum, these results demonstrate the ability of LLMs in accurately predicting question relevance for personas.
This allows us to reliably simulate CTR signals using our proposed user simulator.

\subsection{Recommendation Quality Results}
\label{subsec:ctr_optimization_evaluation}

In the following we present the evaluation of recommendation quality for our approach and ablations.

\subsubsection{E-commerce Question Suggestion}\label{subsubsec:shopping_domain}

\begin{figure*}[ht!]
    \centering
    \includegraphics[width=1\textwidth]{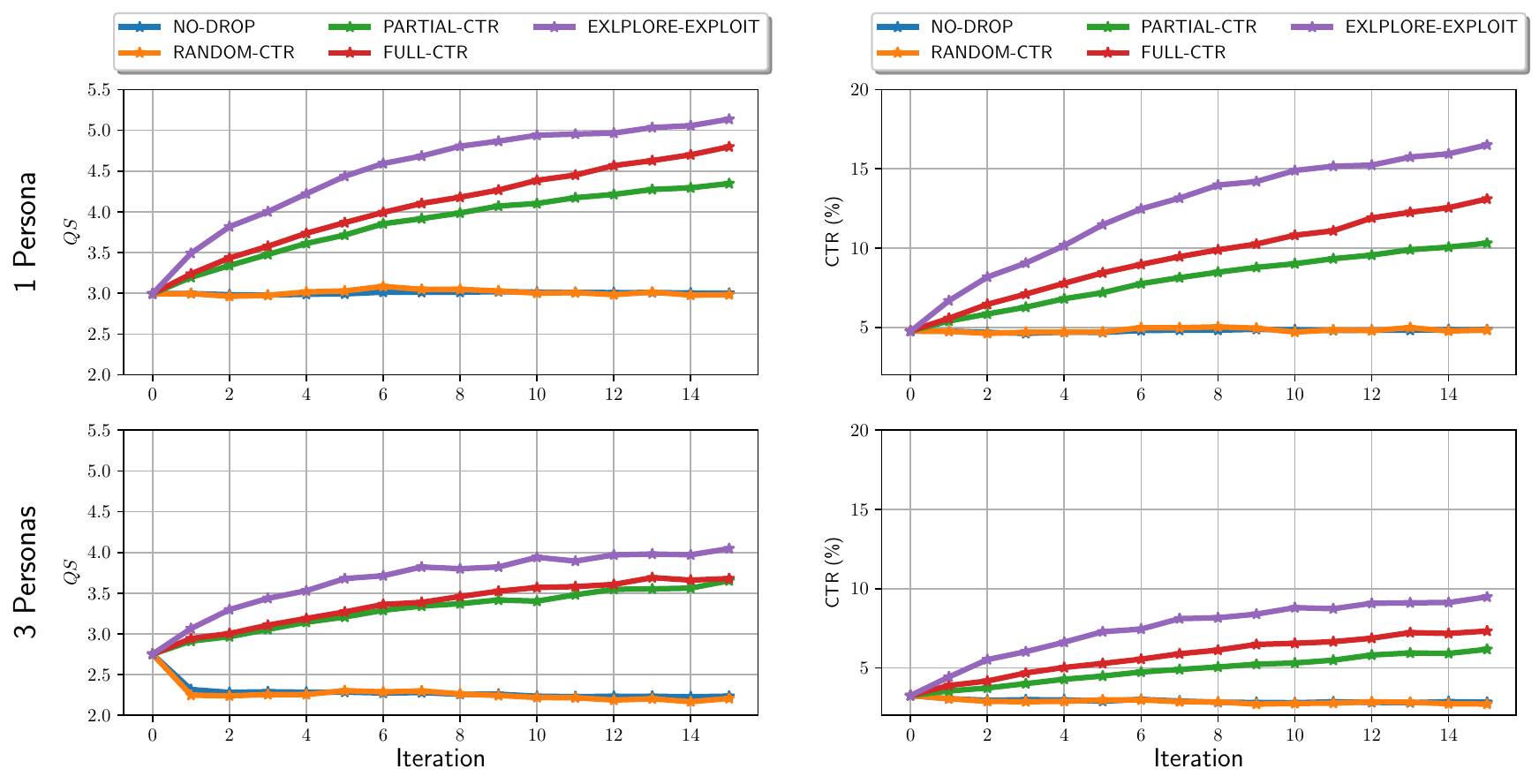}
    \caption{The plots on the left hand side show the average question scores, while the right hand side shows the \ctr scores for the e-commerce domain for personas with 1 and 3 preferences. For personas with a single preference, the results are averaged across 5 different personas (see Figure \ref{fig:shopping_single_preference}.) 
    }
    \label{fig:shopping_mixed_pref}
\end{figure*}

Figure \ref{fig:shopping_mixed_pref} shows the averaged results obtained across different personas (with varying preference counts) for all approaches.

\paragraph{Exploration is key for improving generated recommendations.}
Overall, \ee achieves a statistically significantly higher \ctr when compared against all other competing approaches.\footnote{\textit{p}-value < .001, as measured by the Z-test for proportions.} This shows \ee utilizes LLMs to both explore new questions, which ensures that they are relevant for multiple  personas, and at the same time, whenever \ctr improvement is observed through our user simulations, it can pivot and exploit such question shapes and aspects to generate questions that are likely to be relevant for the underlying population base.
The results clearly demonstrate the importance of not only exploiting identified high-CTR items, but also actively exploring to discover the hidden preferences of the target population, which the \ee is method is able to do.

\begin{figure*}[ht!]
    \centering
    \includegraphics[width=1\textwidth]{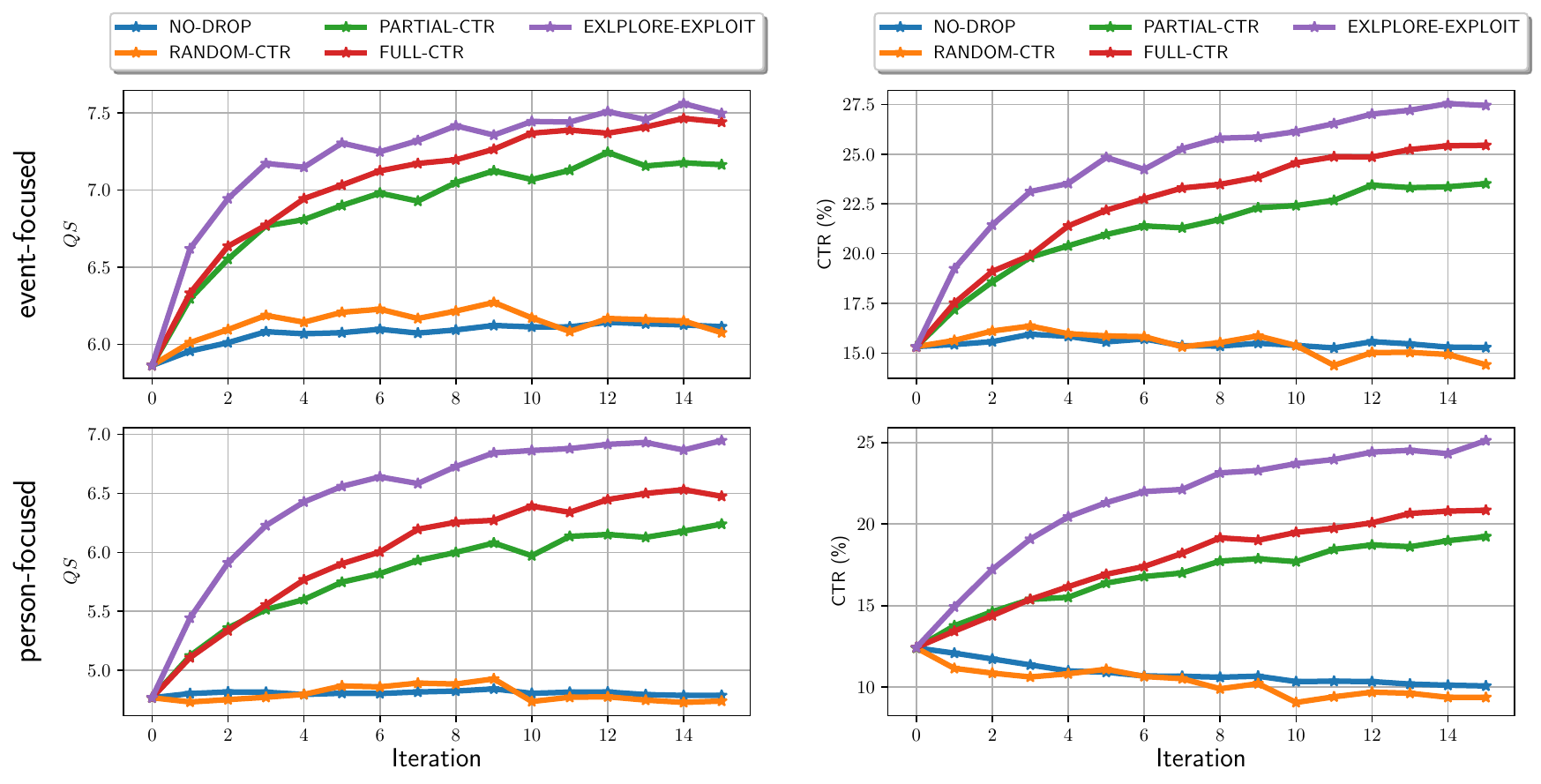}
    \caption{Average question scores and CTRs for the \partialctr, \ctr and \ee methods on general knowledge domain for personas with a single preference.}
    \label{fig:wiki_single_focus}
\end{figure*}

\noindent\textbf{Question Relevance scores improve iteratively.} \ee demonstrates a nearly constant increase in scores. If we compare \QPI{0} against \QPI{15}, we see an improvement of more than $\Delta=+2$ points. 
We also observe that \fullctr performs better than \partialctr, however the gap is relatively small. 
This indicates that the LLM is able to make use of the \ctr signal provided via in-context learning (e.g. question and corresponding \ctr score), without having an explicit strategy to explore new questions and exploit the best performing ones, however, its effectiveness is limited.
We also more explicitly investigated the LLM's optimization capability through the \ctr signal using a synthetic setup, where we scored questions based on their length, allowing us to have a clear and deterministic signal. This further validated our conclusions. For a more detailed analysis of the results for this setup  we refer the reader to \Cref{appendix:add_eval}.

\paragraph{Using observed CTR is critical for Exploitation.} 
The large gap between \partialctr and the \nodrop and \randomdrop baselines shows the importance of using the \ctr signal to drop the worst performing questions from the question pool.
We also note that \randomdrop and \nodrop baselines perform equally poor and fail to improve the question relevance scores across iterations. This is expected since they do not make use of the \ctr signal to generate more engaging questions or filter out the worst performing ones.

\noindent\textbf{\ctr consistently improves in all settings.} Since \ctr is simulated by relying on question relevance score, here too, the best performing approach is \ee. From \QPI{0} to \QPI{15} we notice an improvement of more than $\Delta\,\ctr=+11\%$ percentage points for populations with single personas, and more than $\Delta\,\ctr=+7\%$ for populations with 3 personas. 
A similar trend, as for question relevance, is observed between \fullctr and \partialctr, where the gap in their \ctr scores is at most 3\%. When we compare \ee against \fullctr and \partialctr, we observe that the differences in the question relevance scores are further amplified for the \ctr where \ee obtains significantly higher scores.
Moreover, \ctr results from \nodrop and \randomdrop highlight the importance of the feedback loop, allowing LLMs to iteratively refine the questions in \QP, thus, increasing their relevance and thereby their \ctr. 

\paragraph{Example Recommendations:} To show how \ee leverages CTR signals, we provide examples from the first and last iterations of the model in \Cref{sec:examples}.
We observe that our \ee approach can effectively discover the hidden preferences of the user population.
For example, if the simulated users for the query \example{spray bottles} have a hidden preference for \texttt{Ethical Considerations}, our approach converges to generate questions such as \example{Are there eco-friendly biodegradable options for spray bottles?} without any direct knowledge of the user preferences. 
Similarly, if the user population includes a preference for \texttt{Quality}, optimizing questions for the query \example{cookware sets} with our approach results in highly relevant questions such as \example{Are copper cookware sets prone to tarnishing over time?}.

\subsubsection{General Knowledge Domain}
Figure \ref{fig:wiki_single_focus} shows the results for the general knowledge domain (see Figure \ref{fig:wiki_single_focus_all} for the results for all personas). Here too, as in the e-commerce domain, approaches that make use of \ctr are able to iteratively improve question relevance and \ctr, while the \randomdrop and \nodrop baselines fail to improve.
These results demonstrate that our approach is applicable across domains and tasks.

\subsection{Persona-level Results}
Evaluations for all personas are listed in \Cref{sec:AllResults}. Results show that our approach can discover the preferences of a diverse set of personas.

\subsection{Human Evaluation of Recommendations}
\label{subsubsec:human_qualitative_eval}

For the best performing \ee approach, we carried out a human evaluation to understand the preferences of human annotators for the item pool, \QP. Annotators, without knowing the source iteration of the questions, performed a pairwise preference annotation of \QPI{0} against \QPI{15} for the e-commerce domain. 

Out of 25 pairwise comparisons, in 88\% (22 cases), annotators judged \QPI{15} as their preferred question set for a given persona and topic. This result further validates the improvements we see in terms of question relevance scores as well as \ctr.

\section{Considerations for Online Deployment}
In this work we used a simulator to facilitate offline development. This approach allowed us to develop and test various algorithms without needing to directly involve real users or incur the associated costs and risks. While having the simulator  as a stand-in for real user feedback served to significantly expedite the development process, transitioning to real-world deployment presents a set of new challenges and requires modifications.

The core requirement is to replace the simulator-generated feedback with actual user engagement. This means the recommendations generated by the system must be served to real users, and their interactions (such as \ctr and dwell time) need to be collected and stored for each input context. As real engagement is noisy and sparse, implementing efficient data pipelines is key. Caching might be needed to store and serve the item pool for each context in order to minimize latency.
Furthermore, the optimization iterations may need to run at fixed intervals, e.g. after every \textit{N} user engagements or based on a predetermined time schedule. This periodic and continuous optimization ensures that the model evolves in response to the latest user data. 

Finally, moving from the cohort-based personalization  explored here (where recommendations rely on generalized user segments) to user-level personalization will require further changes. Tailoring recommendations to individuals will require collecting more granular data.

\section{Conclusion}
\label{sec:conclusions}

We proposed a novel method to improve LLM-based generative recommender systems by iteratively refining recommendations based on implicit feedback loops from \ctr signals.
We additionally defined a user simulator to effectively simulate user interactions with such recommended items.

Our novel Generative \ee approach does not require any fine-tuning, and only relies on an LLM optimizer using in-context learning by synthesizing observed CTR performance data and incorporating them into the prompt.
Experiments with our approach show that while leveraging historical CTR data is crucial to \textit{exploit} known engagement patterns, the inclusion of a generative \textit{explore} phase is equally important for discovering user preferences.
Evaluations on the task of question generation across two domains (e-commerce and world knowledge) show that our proposed generative recommender approach is able to generate questions that are highly relevant to its user population in just a few iterations, which in turn results in higher engagement as measured through CTR.

The generative \ee method is is particularly suitable for tasks where there are many potentially valid suggestions that can be generated.
While we studied question generation in this paper, it can applied to a range of tasks such as summarization \cite{fetahu-etal-2023-instructpts}, personalized headline generations \cite{cai-etal-2023-generating}, and follow-up question suggestion \cite{fetahu-etal-2023-follow}. 
By avoiding the use of reward models or fine-tuning, our approach can effectively scale to scenarios with billions of items, while also being able to support user-level personalization.

\section*{Limitations}

There are some limitations in the current version of this work that we would like to highlight. 

\paragraph{User-level Personalization.}
In this work we did not model individual users, but instead modeled cohorts of users. Our current approach can be adapted to address this gap by extending the feedback loop to be user specific. The advantages of our training-free approach are even more relevant in such a setting, allowing us to efficiently scale personalized recommendations to millions of users.
We leave this line of inquiry for future work.

\paragraph{Budget Constraints.} Due to limitations on our budget we could not run all possible experiments.
For example, our work presents only the question generation task, while, in principle, the proposed framework can be applied to different recommendation tasks. Similarly, we couldn't host open source LLMs of quality close enough to ChatGPT (i.e., with size >70B): in preliminary experiments we noticed that smaller models are not great in following our instructions and to reason over the numerical CTR signal.

\paragraph{Offline Experimental Framework.} As we don't have a real system to rely on, it is impossible to run real user studies. For this reason, we report only experiments with simulated users/persona. We tried our best in assessing the quality of the simulator and we believe the results are acceptable.

\paragraph{Baselines.}
We did not adapt existing optimization approaches, e.g., multi-armed bandits, to our task. While this is in principle possible by generating a very large pool of questions, this will remain a static set; in our approach the questions pool is dynamic so we believe a comparison would not be fully fair. We did not consider the possibility of a hybrid approach, which we leave as future work.

\section*{Acknowledgements}
We would like to thank Eugene Agichtein, Oleg Rokhlenko, and Saar Kuzi for their feedback.

\bibliography{main}

\begin{thebibliography}{41}
\expandafter\ifx\csname natexlab\endcsname\relax\def\natexlab#1{#1}\fi

\bibitem[{Afsar et~al.(2023)Afsar, Crump, and
  Far}]{DBLP:journals/csur/AfsarCF23}
Mohammad~Mehdi Afsar, Trafford Crump, and Behrouz~H. Far. 2023.
\newblock \href {https://doi.org/10.1145/3543846} {Reinforcement learning based
  recommender systems: {A} survey}.
\newblock \emph{{ACM} Comput. Surv.}, 55(7):145:1--145:38.

\bibitem[{Cai et~al.(2023)Cai, Song, Cho, Wang, Wang, Yu, Liu, and
  Yu}]{cai-etal-2023-generating}
Pengshan Cai, Kaiqiang Song, Sangwoo Cho, Hongwei Wang, Xiaoyang Wang, Hong Yu,
  Fei Liu, and Dong Yu. 2023.
\newblock \href {https://doi.org/10.18653/v1/2023.acl-long.183} {Generating
  user-engaging news headlines}.
\newblock In \emph{Proceedings of the 61st Annual Meeting of the Association
  for Computational Linguistics (Volume 1: Long Papers)}, pages 3265--3280,
  Toronto, Canada. Association for Computational Linguistics.

\bibitem[{Carmel et~al.(2020)Carmel, Haramaty, Lazerson, Lewin-Eytan, and
  Maarek}]{Carmel2020}
David Carmel, Elad Haramaty, Arnon Lazerson, Liane Lewin-Eytan, and Yoelle
  Maarek. 2020.
\newblock \href
  {https://www.amazon.science/publications/why-do-people-buy-irrelevant-items-in-voice-product-search}
  {Why do people buy seemingly irrelevant items in voice product search?}
\newblock In \emph{WSDM 2020}.

\bibitem[{Dong et~al.(2023)Dong, Chen, Liu, Polak, and
  Zhang}]{dong2023musechat}
Zhikang Dong, Bin Chen, Xiulong Liu, Pawel Polak, and Peng Zhang. 2023.
\newblock Musechat: A conversational music recommendation system for videos.
\newblock \emph{arXiv preprint arXiv:2310.06282}.

\bibitem[{Dupret and Piwowarski(2008)}]{dupret2008UBM}
Georges~E Dupret and Benjamin Piwowarski. 2008.
\newblock A user browsing model to predict search engine click data from past
  observations.
\newblock In \emph{Proceedings of the 31st annual international ACM SIGIR
  conference on Research and development in information retrieval}, pages
  331--338.

\bibitem[{Fetahu et~al.(2023{\natexlab{a}})Fetahu, Chen, Rokhlenko, and
  Malmasi}]{fetahu-etal-2023-instructpts}
Besnik Fetahu, Zhiyu Chen, Oleg Rokhlenko, and Shervin Malmasi.
  2023{\natexlab{a}}.
\newblock \href {https://doi.org/10.18653/v1/2023.emnlp-industry.63}
  {{I}nstruct{PTS}: Instruction-tuning {LLM}s for product title summarization}.
\newblock In \emph{Proceedings of the 2023 Conference on Empirical Methods in
  Natural Language Processing: Industry Track}, pages 663--674, Singapore.
  Association for Computational Linguistics.

\bibitem[{Fetahu et~al.(2023{\natexlab{b}})Fetahu, Faustini, Fang, Castellucci,
  Rokhlenko, and Malmasi}]{fetahu-etal-2023-follow}
Besnik Fetahu, Pedro Faustini, Anjie Fang, Giuseppe Castellucci, Oleg
  Rokhlenko, and Shervin Malmasi. 2023{\natexlab{b}}.
\newblock \href {https://doi.org/10.18653/v1/2023.findings-emnlp.24} {Follow-on
  question suggestion via voice hints for voice assistants}.
\newblock In \emph{Findings of the Association for Computational Linguistics:
  EMNLP 2023}, pages 310--325, Singapore. Association for Computational
  Linguistics.

\bibitem[{Fu et~al.(2023)Fu, Li, Wu, Wang, Dong, Zhao, Zhao, Guo, and
  Tang}]{fu2023unified}
Zichuan Fu, Xiangyang Li, Chuhan Wu, Yichao Wang, Kuicai Dong, Xiangyu Zhao,
  Mengchen Zhao, Huifeng Guo, and Ruiming Tang. 2023.
\newblock \href {http://arxiv.org/abs/2312.10743} {A unified framework for
  multi-domain ctr prediction via large language models}.

\bibitem[{Ghanem et~al.(2022)Ghanem, Lutz~Coleman, Rivard~Dexter, von~der Ohe,
  and Fyshe}]{ghanem-etal-2022-question}
Bilal Ghanem, Lauren Lutz~Coleman, Julia Rivard~Dexter, Spencer von~der Ohe,
  and Alona Fyshe. 2022.
\newblock \href {https://doi.org/10.18653/v1/2022.findings-acl.168} {Question
  generation for reading comprehension assessment by modeling how and what to
  ask}.
\newblock In \emph{Findings of the Association for Computational Linguistics:
  ACL 2022}, pages 2131--2146, Dublin, Ireland. Association for Computational
  Linguistics.

\bibitem[{Haramaty et~al.(2023)Haramaty, Karnin, Lazerson, Lewin-Eytan, and
  Maarek}]{10.1145/3604915.3608836}
Elad Haramaty, Zohar Karnin, Arnon Lazerson, Liane Lewin-Eytan, and Yoelle
  Maarek. 2023.
\newblock \href {https://doi.org/10.1145/3604915.3608836} {Extended conversion:
  Capturing successful interactions in voice shopping}.
\newblock In \emph{Proceedings of the 17th ACM Conference on Recommender
  Systems}, RecSys '23, page 826–832, New York, NY, USA. Association for
  Computing Machinery.

\bibitem[{Huber et~al.(2022)Huber, Aghajanyan, Oguz, Okhonko, Yih, Gupta, and
  Chen}]{huber-etal-2022-ccqa}
Patrick Huber, Armen Aghajanyan, Barlas Oguz, Dmytro Okhonko, Scott Yih, Sonal
  Gupta, and Xilun Chen. 2022.
\newblock \href {https://doi.org/10.18653/v1/2022.findings-naacl.184} {{CCQA}:
  A new web-scale question answering dataset for model pre-training}.
\newblock In \emph{Findings of the Association for Computational Linguistics:
  NAACL 2022}, pages 2402--2420, Seattle, United States. Association for
  Computational Linguistics.

\bibitem[{Kostric et~al.(2021)Kostric, Balog, and
  Radlinski}]{kostric2021soliciting}
Ivica Kostric, Krisztian Balog, and Filip Radlinski. 2021.
\newblock Soliciting user preferences in conversational recommender systems via
  usage-related questions.
\newblock In \emph{Proceedings of the 15th ACM Conference on Recommender
  Systems}, pages 724--729.

\bibitem[{Li et~al.(2023{\natexlab{a}})Li, Zhang, Wang, Xiong, Lu, and
  Medioni}]{li2023gpt4rec}
Jinming Li, Wentao Zhang, Tian Wang, Guanglei Xiong, Alan Lu, and Gerard
  Medioni. 2023{\natexlab{a}}.
\newblock \href {http://arxiv.org/abs/2304.03879} {Gpt4rec: A generative
  framework for personalized recommendation and user interests interpretation}.

\bibitem[{Li et~al.(2023{\natexlab{b}})Li, Zhang, Liu, and Chen}]{li2023large}
Lei Li, Yongfeng Zhang, Dugang Liu, and Li~Chen. 2023{\natexlab{b}}.
\newblock \href {http://arxiv.org/abs/2309.01157} {Large language models for
  generative recommendation: A survey and visionary discussions}.

\bibitem[{Li et~al.(2023{\natexlab{c}})Li, Wang, Chi, and Chen}]{li2023prompt}
Pan Li, Yuyan Wang, Ed~H Chi, and Minmin Chen. 2023{\natexlab{c}}.
\newblock Prompt tuning large language models on personalized aspect extraction
  for recommendations.
\newblock \emph{arXiv preprint arXiv:2306.01475}.

\bibitem[{Li et~al.(2023{\natexlab{d}})Li, Chen, Hou, and Tang}]{li2023ctrl}
Xiangyang Li, Bo~Chen, Lu~Hou, and Ruiming Tang. 2023{\natexlab{d}}.
\newblock \href {http://arxiv.org/abs/2306.02841} {Ctrl: Connect collaborative
  and language model for ctr prediction}.

\bibitem[{Lin et~al.(2023)Lin, Dai, Xi, Liu, Chen, Li, Zhu, Guo, Yu, Tang
  et~al.}]{lin2023can}
Jianghao Lin, Xinyi Dai, Yunjia Xi, Weiwen Liu, Bo~Chen, Xiangyang Li, Chenxu
  Zhu, Huifeng Guo, Yong Yu, Ruiming Tang, et~al. 2023.
\newblock How can recommender systems benefit from large language models: A
  survey.
\newblock \emph{arXiv preprint arXiv:2306.05817}.

\bibitem[{Liu et~al.(2023)Liu, Liu, Zhou, Lv, Zhou, and Zhang}]{liu2023chatgpt}
Junling Liu, Chao Liu, Peilin Zhou, Renjie Lv, Kang Zhou, and Yan Zhang. 2023.
\newblock \href {http://arxiv.org/abs/2304.10149} {Is chatgpt a good
  recommender? a preliminary study}.

\bibitem[{Metzler et~al.(2021)Metzler, Tay, Bahri, and
  Najork}]{metzler2021rethinking}
Donald Metzler, Yi~Tay, Dara Bahri, and Marc Najork. 2021.
\newblock Rethinking search: making domain experts out of dilettantes.
\newblock In \emph{Acm sigir forum}, volume~55, pages 1--27. ACM New York, NY,
  USA.

\bibitem[{Mitra et~al.(2021)Mitra, Jain, Veerubhotla, and
  Gupta}]{DBLP:conf/kdd/MitraJVG21}
Rajarshee Mitra, Rhea Jain, Aditya~Srikanth Veerubhotla, and Manish Gupta.
  2021.
\newblock \href {https://doi.org/10.1145/3447548.3469403} {Zero-shot
  multi-lingual interrogative question generation for "people also ask" at
  bing}.
\newblock In \emph{{KDD} '21: The 27th {ACM} {SIGKDD} Conference on Knowledge
  Discovery and Data Mining, Virtual Event, Singapore, August 14-18, 2021},
  pages 3414--3422. {ACM}.

\bibitem[{Molaei et~al.(2021)Molaei, Havvaei, Zare, and
  Jalili}]{DBLP:journals/access/MolaeiH0J21}
Soheila Molaei, Amirhossein Havvaei, Hadi Zare, and Mahdi Jalili. 2021.
\newblock \href {https://doi.org/10.1109/ACCESS.2021.3054818} {Collaborative
  deep forest learning for recommender systems}.
\newblock \emph{{IEEE} Access}, 9:22053--22061.

\bibitem[{Mulla and Gharpure(2023)}]{mulla2023automatic}
Nikahat Mulla and Prachi Gharpure. 2023.
\newblock Automatic question generation: a review of methodologies, datasets,
  evaluation metrics, and applications.
\newblock \emph{Progress in Artificial Intelligence}, 12(1):1--32.

\bibitem[{Najork(2023)}]{DBLP:conf/sigir/Najork23}
Marc Najork. 2023.
\newblock \href {https://doi.org/10.1145/3539618.3591871} {Generative
  information retrieval}.
\newblock In \emph{Proceedings of the 46th International {ACM} {SIGIR}
  Conference on Research and Development in Information Retrieval, {SIGIR}
  2023, Taipei, Taiwan, July 23-27, 2023}, page~1. {ACM}.

\bibitem[{Ni et~al.(2019)Ni, Li, and McAuley}]{ni2019trec}
Jianmo Ni, Jiacheng Li, and Julian McAuley. 2019.
\newblock Justifying recommendations using distantly-labeled reviews and
  fine-grained aspects.
\newblock In \emph{Proceedings of the 2019 conference on empirical methods in
  natural language processing and the 9th international joint conference on
  natural language processing (EMNLP-IJCNLP)}, pages 188--197.

\bibitem[{OpenAI(2023)}]{openai2023gpt4}
OpenAI. 2023.
\newblock \href {http://arxiv.org/abs/2303.08774} {Gpt-4 technical report}.

\bibitem[{Pothirattanachaikul et~al.(2020)Pothirattanachaikul, Yamamoto,
  Yamamoto, and Yoshikawa}]{10.1145/3372923.3404786}
Suppanut Pothirattanachaikul, Takehiro Yamamoto, Yusuke Yamamoto, and Masatoshi
  Yoshikawa. 2020.
\newblock \href {https://doi.org/10.1145/3372923.3404786} {Analyzing the
  effects of "people also ask" on search behaviors and beliefs}.
\newblock In \emph{Proceedings of the 31st ACM Conference on Hypertext and
  Social Media}, HT '20, page 101–110, New York, NY, USA. Association for
  Computing Machinery.

\bibitem[{Qin et~al.(2023)Qin, Cai, Jin, Yan, Liang, Zhu, Lin, Han, Ding, Wang,
  Xie, Qi, Liu, Sun, and Zhou}]{qin-etal-2023-webcpm}
Yujia Qin, Zihan Cai, Dian Jin, Lan Yan, Shihao Liang, Kunlun Zhu, Yankai Lin,
  Xu~Han, Ning Ding, Huadong Wang, Ruobing Xie, Fanchao Qi, Zhiyuan Liu,
  Maosong Sun, and Jie Zhou. 2023.
\newblock \href {https://doi.org/10.18653/v1/2023.acl-long.499} {{W}eb{CPM}:
  Interactive web search for {C}hinese long-form question answering}.
\newblock In \emph{Proceedings of the 61st Annual Meeting of the Association
  for Computational Linguistics (Volume 1: Long Papers)}, pages 8968--8988,
  Toronto, Canada. Association for Computational Linguistics.

\bibitem[{Rajput et~al.(2023)Rajput, Mehta, Singh, Keshavan, Vu, Heldt, Hong,
  Tay, Tran, Samost et~al.}]{rajput2023recommender}
Shashank Rajput, Nikhil Mehta, Anima Singh, Raghunandan~H Keshavan, Trung Vu,
  Lukasz Heldt, Lichan Hong, Yi~Tay, Vinh~Q Tran, Jonah Samost, et~al. 2023.
\newblock Recommender systems with generative retrieval.
\newblock \emph{arXiv preprint arXiv:2305.05065}.

\bibitem[{Sarwar et~al.(2002)Sarwar, Karypis, Konstan, and
  Riedl}]{sarwar2002incremental}
Badrul Sarwar, George Karypis, Joseph Konstan, and John Riedl. 2002.
\newblock Incremental singular value decomposition algorithms for highly
  scalable recommender systems.
\newblock In \emph{Fifth international conference on computer and information
  science}, volume~1, pages 27--8. Citeseer.

\bibitem[{Tang et~al.(2023)Tang, Huan, Li, Zhang, Hu, Fu, Zhou, and
  Li}]{tang2023one}
Zuoli Tang, Zhaoxin Huan, Zihao Li, Xiaolu Zhang, Jun Hu, Chilin Fu, Jun Zhou,
  and Chenliang Li. 2023.
\newblock One model for all: Large language models are domain-agnostic
  recommendation systems.
\newblock \emph{arXiv preprint arXiv:2310.14304}.

\bibitem[{Vedula et~al.(2024)Vedula, Rokhlenko, and
  Malmasi}]{vedula2024question}
Nikhita Vedula, Oleg Rokhlenko, and Shervin Malmasi. 2024.
\newblock Question suggestion for conversational shopping assistants using
  product metadata.
\newblock \emph{arXiv preprint arXiv:2405.01738}.

\bibitem[{Wang et~al.(2023{\natexlab{a}})Wang, Lin, Feng, He, and
  Chua}]{wang2023generative}
Wenjie Wang, Xinyu Lin, Fuli Feng, Xiangnan He, and Tat-Seng Chua.
  2023{\natexlab{a}}.
\newblock \href {http://arxiv.org/abs/2304.03516} {Generative recommendation:
  Towards next-generation recommender paradigm}.

\bibitem[{Wang et~al.(2023{\natexlab{b}})Wang, Tang, Zhao, Wang, and
  Wen}]{wang-etal-2023-rethinking-evaluation}
Xiaolei Wang, Xinyu Tang, Xin Zhao, Jingyuan Wang, and Ji-Rong Wen.
  2023{\natexlab{b}}.
\newblock \href {https://doi.org/10.18653/v1/2023.emnlp-main.621} {Rethinking
  the evaluation for conversational recommendation in the era of large language
  models}.
\newblock In \emph{Proceedings of the 2023 Conference on Empirical Methods in
  Natural Language Processing}, pages 10052--10065, Singapore. Association for
  Computational Linguistics.

\bibitem[{Wu et~al.(2023)Wu, Liu, Hu, Fan, Liu, Li, Wu, and
  Tang}]{wu2023leveraging}
Jiahao Wu, Qijiong Liu, Hengchang Hu, Wenqi Fan, Shengcai Liu, Qing Li,
  Xiao-Ming Wu, and Ke~Tang. 2023.
\newblock Leveraging large language models (llms) to empower training-free
  dataset condensation for content-based recommendation.
\newblock \emph{arXiv preprint arXiv:2310.09874}.

\bibitem[{Xi et~al.(2023)Xi, Liu, Lin, Zhu, Chen, Tang, Zhang, Zhang, and
  Yu}]{xi2023towards}
Yunjia Xi, Weiwen Liu, Jianghao Lin, Jieming Zhu, Bo~Chen, Ruiming Tang, Weinan
  Zhang, Rui Zhang, and Yong Yu. 2023.
\newblock Towards open-world recommendation with knowledge augmentation from
  large language models.
\newblock \emph{arXiv preprint arXiv:2306.10933}.

\bibitem[{Yang et~al.(2021)Yang, Han, Li, Zuo, and Yu}]{yang2021improving}
Bowen Yang, Cong Han, Yu~Li, Lei Zuo, and Zhou Yu. 2021.
\newblock Improving conversational recommendation systems' quality with
  context-aware item meta information.
\newblock \emph{arXiv preprint arXiv:2112.08140}.

\bibitem[{Yang et~al.(2023)Yang, Wang, Lu, Liu, Le, Zhou, and
  Chen}]{yang2023LLMsOptimizers}
Chengrun Yang, Xuezhi Wang, Yifeng Lu, Hanxiao Liu, Quoc~V. Le, Denny Zhou, and
  Xinyun Chen. 2023.
\newblock \href {http://arxiv.org/abs/2309.03409} {Large language models as
  optimizers}.

\bibitem[{Zhang et~al.(2023{\natexlab{a}})Zhang, Sheng, Chen, Li, Deng, Wang,
  and Chua}]{DBLP:journals/corr/abs-2310-10108}
An~Zhang, Leheng Sheng, Yuxin Chen, Hao Li, Yang Deng, Xiang Wang, and
  Tat{-}Seng Chua. 2023{\natexlab{a}}.
\newblock \href {https://doi.org/10.48550/ARXIV.2310.10108} {On generative
  agents in recommendation}.
\newblock \emph{CoRR}, abs/2310.10108.

\bibitem[{Zhang et~al.(2023{\natexlab{b}})Zhang, Liu, Mao, Ma, Xu, Ma, and
  Tian}]{zhang2023reranking}
Junqi Zhang, Yiqun Liu, Jiaxin Mao, Weizhi Ma, Jiazheng Xu, Shaoping Ma, and
  Qi~Tian. 2023{\natexlab{b}}.
\newblock User behavior simulation for search result re-ranking.
\newblock \emph{ACM Transactions on Information Systems}, 41(1):1--35.

\bibitem[{Zhang et~al.(2023{\natexlab{c}})Zhang, Feng, Zhang, Bao, Wang, and
  He}]{zhang2023collm}
Yang Zhang, Fuli Feng, Jizhi Zhang, Keqin Bao, Qifan Wang, and Xiangnan He.
  2023{\natexlab{c}}.
\newblock Collm: Integrating collaborative embeddings into large language
  models for recommendation.
\newblock \emph{arXiv preprint arXiv:2310.19488}.

\bibitem[{Zheng et~al.(2023)Zheng, Qiu, Hu, Wu, Zhu, and
  Xiong}]{zheng2023generative}
Zhi Zheng, Zhaopeng Qiu, Xiao Hu, Likang Wu, Hengshu Zhu, and Hui Xiong. 2023.
\newblock \href {http://arxiv.org/abs/2307.02157} {Generative job
  recommendations with large language model}.

\end{thebibliography}

\appendix
\clearpage

\onecolumn

\section*{\centering Appendix}

\section{Details of Dataset Topics}
\label{sec:input_topics}

Table~\ref{tab:input_topics} lists the 50 product categories and the 50 Wikipedia article titles used as input topics in this work.

\begin{table*}[!bh]
    \centering
    \resizebox{\textwidth}{!}{\begin{tabular}{ll|ll}
    \toprule
    \multicolumn{2}{c}{\textbf{E-commerce}} & \multicolumn{2}{c}{\textbf{Wikipedia}} \\
    \midrule
    Spray Bottles & Home Bar Furniture & Stoicism & Tabata training \\
    Cookware Sets & Lighting \& Ceiling Fans & Friedrich Nietzsche & Chernobyl disaster \\
    TV Antennas & Vehicle Backup Cameras & Artificial Superintelligence & Chimamanda Ngozi Adichie \\
    DVI & Drills & The Beatles' rooftop concert & Banksy \\
    Measuring Tools \& Scales & Tablet Accessories & Carl Jung & Kabuki \\
    Coaxial Cables & Champagne Glasses & History of film & Surrealist Manifesto \\
    Area Rugs & Window Treatments & Vinson Massif & Great Barrier Reef \\
    Lightning Cables & Diamond Blades & Socotra & Lake Baikal \\
    Kitchen Sinks & Hair Combs & Petra & Avenue of the Baobabs \\
    Wall Plates & Bath Bombs & Ketogenic diet & Mindfulness-based stress reduction \\
    Clips & Table Saw Accessories & Health benefits of pomegranate & Blue zones \\
    Hair Treatment Oils & Speaker & Neuroplasticity & Sinking of the RMS Titanic \\
    Temporary Tattoos & Item Finders & Emancipation Proclamation & The Black Death \\
    Spoons & Computer Cases & Fall of the Berlin Wall & Rosetta Stone discovery \\
    Boot \& Shoe Covers & Racks, Shelves \& Drawers & Beekeeping & Parkour \\
    Fuses & Surveillance Video Recorders & Speedcubing & Citizen science \\
    Computers \& Accessories & Over-Ear Headphones & Flash mob & Sand sculpture \\
    Wireless Access Points & Garage Storage & Gödel's incompleteness theorems & The Banach–Tarski paradox \\
    Safety Work Gloves & Refillable Containers & Poincaré conjecture & Ramanujan's lost notebook \\
    Tea Accessories & Camera \& Photo & Boolean algebra & Fermat's Last Theorem \\
    Bathroom Vanities & Specialty Tools \& Gadgets & Periodic table & Schrödinger's cat \\
    Bookshelf Albums & Lash Enhancers \& Primers & Great Oxygenation Event & Dark matter \\
    Telescopes & Conditioners & Plate tectonics & Bioluminescence \\
    Dining Chair Slipcovers & Electrical & Diogenes of Sinope & Leonardo da Vinci \\
    Single Rods & Vacuums & Malala Yousafzai & Marie Curie \\
    \bottomrule
    \end{tabular}}
    \caption{List of the 50 product categories and 50 Wikipedia articles that are used as the input topics.}
    \label{tab:input_topics}
\end{table*}

\section{Wikipedia Data}
\label{sec:datasets_appx}

For the Wikipedia domain, we obtain a diverse set of 72 Wikipedia article titles (These generated titles were verified to be actual Wikipedia pages)
by prompting \llm to write 6 diverse and interesting Wikipedia article titles for each of the 12 Wikipedia categories\footnote{\url{https://en.wikipedia.org/wiki/Wikipedia:Contents/Categories}, general reference category is excluded} and then randomly sample 50 articles.

\section{User Simulator Details}
\label{sec:simulator_details}

\subsection{Effect of Softmax Temperature}\label{sec:SoftmaxTemp}
\begin{figure}[!htb]
    \centering
    \includegraphics[width=0.7\columnwidth]{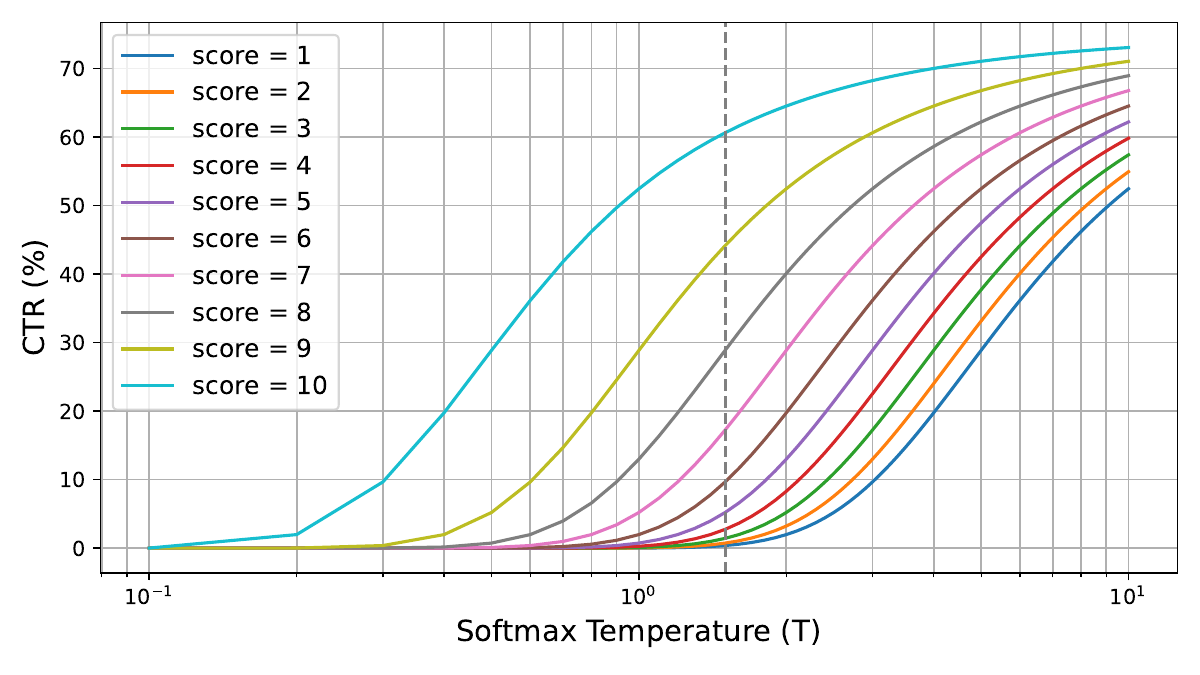}
    \caption{Theoretical CTR values with $RS=11$ for varying temperature and for 3 shown questions ($K=3$) with equal scores ranging from 1 to 10. Vertical dashed line at $T=1.5$ shows the temperature that we use in our simulations.}
    \label{fig:T}
\end{figure}

We wanted to ensure that the user simulator converted the LLM generated relevance scores into meaningfully different simulated user behavior.
However, with a temperature setting of 1 in Equation~\ref{eq:softmax}, the action distribution is always extremely peaked, leading to only extremely relevant questions receiving meaningful click probabilities.
Figure \ref{fig:T} shows the effect of $T$ $K=3, RS=11$, for question scores ranging from 1 to 10, and makes it clear that a higher temperature is to produce diverse action distributions.
Based on this, we set the temperature for our experiments to $T=1.5$, leading to more sensitivity to changes in question relevance score, even when the scores are lower.

\subsection{Click Simulation Analysis} \label{sec:SimCountAnalysis}
\begin{figure*}[!htb]
    \centering
    \includegraphics[width=\textwidth]{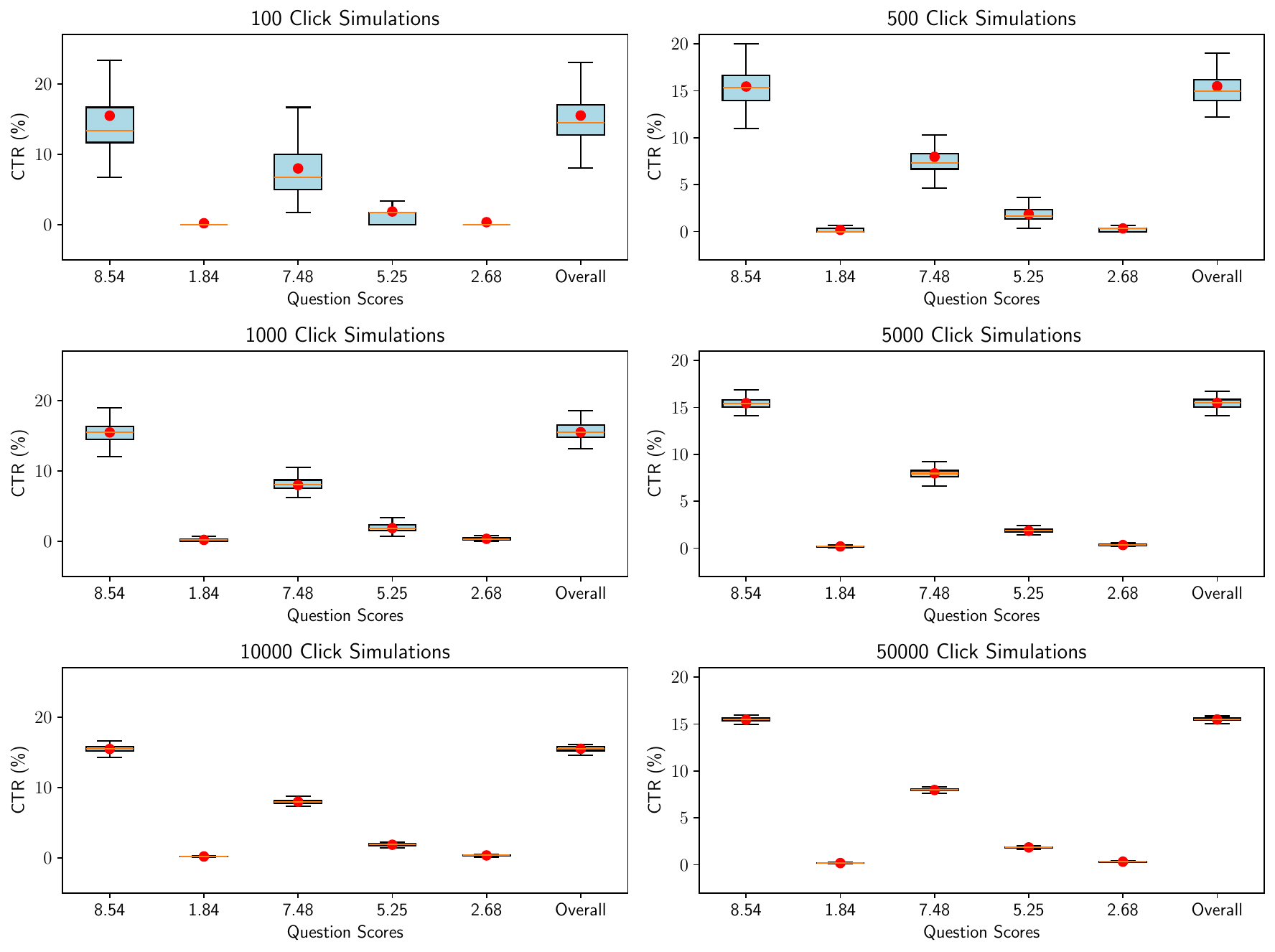}
    \caption{Variation of the CTR for various number of click simulations (S) for a question pool of size 5 with randomly generated question scores. For each S, the simulation is ran 100 times and variance of the resulting CTR values for each question, as well as the overall CTR are shown.}
    \label{fig:ctr_for_s}
\end{figure*}

In real user studies, one only obtains a noisy estimate of user interest through CTR, which is why we draw samples from an action distribution rather than directly reporting the ``true'' CTR of the user population.
Here, we investigate the effect of the number of simulated user interactions (S) in each iteration on the reliability of the resulting CTR values by varying S between 100 and 50,000.
Figure \ref{fig:ctr_for_s} shows the variance in the resulting CTR for the tested S values.
As the number of simulations increase, amount of variation in the obtained CTR values goes down and approaches to the theoretical values.
In this study, we use 5,000 simulations which is both realistic and adds a manageable noise to the CTR calculation process.

\clearpage

\section{Prompts}
\label{sec:Prompts}

\begin{table*}
    \centering
    \begin{tabular}{l|p{0.7\textwidth}}
    \toprule
    Persona & Prompt \\
    \midrule
    Price & The cost of the product is one of the primary considerations. It includes not only the initial price but also long-term costs such as maintenance, operation, or subscription fees \\
    Quality & Customers look at the materials, construction, durability, and overall finish of the product. High quality often correlates with longer lifespan and better performance \\
    Brand Reputation & Well-known brands often carry a perceived assurance of quality, trust, and status. Customers may prefer products from brands with a strong reputation or positive previous experiences \\
    Features \& Functionality & The capabilities of the product, including its features, usability, and whether it meets the customer's needs and expectations, are crucial \\
    Ethical Considerations & Increasingly, customers think about the ethical implications of their purchases, such as sustainability, environmental impact, labor practices, and animal welfare. \\
    \midrule
    Discussion-Focused & This person is more interested in about various arguments on this topic and will be more interested in asking questions that are open-ended and thought-provoking which can lead to further discussions \\
    History-Focused & This person is more interested in learning about the history of the topic and will be more interested in asking questions that are centered around the history of the topic. \\
    Event-Focused & This person is more interested in learning about the events related to the topic and will be more interested in asking questions that are centered around the events related to the topic \\
    Person-Focused & This person is more interested in learning about the people related to the topic and will be more interested in asking questions that are centered around the people related to the topic \\
    Location-Focused & This person is more interested in learning about the locations related to the topic and will be more interested in asking questions that are centered around the locations related to the topic \\
    \bottomrule
    \end{tabular}
    \caption{Personas and their corresponding prompts used for question scoring.}
    \label{tab:Personas}
\end{table*}
\begin{figure*}[t]
    \centering
    \begin{minipage}{0.45\textwidth}
        \centering
    \fbox{\parbox{\textwidth}{%
Judge a given question for its relevance to a given customer's shopping interests. Score the question on a scale from 1 to 10 based on its relevance to the customer's shopping interests.
\\
\\
Use the scoring guide below to score a question for a given customer's shopping interests:
\\
\\
10: extremely relevant \\
8 to 9: very relevant \\
6 to 7: probably relevant \\
4 to 5: may or may not be relevant \\
2 to 3: most likely not relevant \\
1: definitely not relevant \\
\\
Customer's Shopping Interest(s): \\
<persona description>
\\
\\
Question: <question> \\
    }} 
    \end{minipage}
    \hfill
    \begin{minipage}{0.45\textwidth}
        \centering
        \fbox{%
    \parbox{\columnwidth}{%
Judge a given question for how interesting it is to a given person who is looking at a Wikipedia page. Score the question on a scale from 1 to 10 based on how interesting it is to the person.
\\
\\
Use the scoring guide below to score a question for how interesting it is to the given person:
\\
\\
10: extremely interesting \\
8 to 9: very interesting \\
6 to 7: probably interesting \\
4 to 5: may or may not be interesting \\
ķ to 3: most likely not interesting \\
1: definitely not interesting \\
\\
Respond in the following format and write nothing else other than the score:
\\
Score: <score>
\\ 
\\
Article Title: \\
<article title> \\
\\
Person's Background: \\
<persona description> \\
\\
Question: <QUESTION>
\\
    } 
}
    \end{minipage}
    \caption{Prompts used for scoring question relevance to a given simulated persona. Left: E-commerce domain, Right: Wikipedia domain. For E-commerce domain, <persona description> involves the persona (e.g., Quality) followed by its description as given in Table~\ref{tab:Personas}. For Wikipedia domain, <persona description> only involves the corresponding prompt to the persona (e.g., \textit{This person is more interested in ...})}\label{fig:scoring_prompts}
\end{figure*}

There are two sets of prompts required for our experiments: those which define the simulated users, and those which are used to guide an LLM to optimize questions for the simulated population.
Table~\ref{tab:Personas} and Figure~\ref{fig:scoring_prompts} show the prompts used for the simulator.
The persona descriptions from the former are substituted into the prompts from the latter in order to obtain the relevance scores for questions.

For generating and refining the question pool, we use two prompts, again specified per-domain.
Figure~\ref{fig:initial_prompts} shows the prompts used to initialize the question pool for each domain, and Figure~\ref{fig:generation_prompts} shows the prompts used for the \ctr and \ee methods to improve on the question pool based on the measured CTRs.

\begin{figure*}[t]
    \centering
    \begin{minipage}{0.45\textwidth}
        \centering
    \fbox{\parbox{\textwidth}{%
    Write <N> general questions that a person might ask to gain information about '<CATEGORY>'. The questions should be as brief as possible and no more than 15 words. Avoid using words like "best", "where". Make sure that you refer to the category in the questions and the questions are general and grammatical.
    }} 
    \end{minipage}
    \hfill
    \begin{minipage}{0.45\textwidth}
        \centering
        \fbox{%
    \parbox{\textwidth}{%
    Given a Wikipedia article, write <N> short questions a person who is viewing this article might want to ask to quickly learn about some of the information in the article. The questions should be answerable by the information in the article and the goal of the questions is that the person will not need to read the entire article to find the answer. The questions should be no more than 15 words.

    Title: <ARTICLE TITLE>
    } 
}
    \end{minipage}
    \caption{Prompts for generating set of initial questions. Left: E-commerce domain, Right: Wikipedia domain}\label{fig:initial_prompts}
\end{figure*}

\begin{figure*}[t]
    \centering
    \begin{minipage}{0.45\textwidth}
        \centering
    \fbox{\parbox{\textwidth}{%
Your task is writing a new question for the category of Spray Bottles that the customers are likely to ask. Below are the previously written questions for this category and their correspoding click through rates (CTR): 
\\
\\
Question: Can spray bottles be reused? \\
CTR: 0.0\%
\\
\\
Question: How do I properly sterilize a spray bottle for safe reuse? \\
CTR: 0.0\%
\\
\\
\dots
\\
\\
Question: What materials are spray bottles made from? \\
CTR: 13.7\%
\\
\\
Question: How do I choose a spray bottle that won't leak or drip? \\
CTR: 14.0\%
\\
\\
Based on the previous questions and their CTRs, write a novel question that is likely to achieve a high CTR. The question should be grammatical and contain no more than 15 words. Avoid using words like "best", "where". Additionally, the question should not be similar to the previous questions.
\\
\\
Strictly use following format in your response:
\\
New Question: <question>
    }} 
    \end{minipage}
    \hfill
    \begin{minipage}{0.45\textwidth}
        \centering
        \fbox{%
    \parbox{\textwidth}{%
Your task is writing a new question for the category of Spray Bottles that the customers are likely to ask. Below are the previously written questions for this category and their correspoding click through rates (CTR): 
\\
\\
Question: Can spray bottles be reused? \\
CTR: 0.0\%
\\
\\
Question: How do I properly sterilize a spray bottle for safe reuse? \\
CTR: 0.0\%
\\
\\
\dots
\\
\\
Question: What materials are spray bottles made from? \\
CTR: 13.7\%
\\
\\
Question: How do I choose a spray bottle that won't leak or drip? \\
CTR: 14.0\%
\\
\\
Write a novel question that is around the same general topic as the best performing question with highest CTR. If there are already more than 2 questions around that topic, choose the topic of another existing question with a good CTR and write a novel question around that topic. The question should be grammatical and contain no more than 15 words. Avoid using words like "best", "where". Additionally, the question should not be similar to the previous questions.
\\
\\
Strictly use following format in your response:
\\
New Question: <question>
    } 
}
    \end{minipage}
    \caption{Prompts used for generating new questions based on measured CTRs. Left: \fullctr method, Right: \ee method}\label{fig:generation_prompts}
\end{figure*}

\clearpage

\section{Verification of LLM Optimization using Length-based Scoring}
\label{appendix:add_eval}

\begin{figure*}[!htb]
    \centering
    \includegraphics[width=\textwidth]{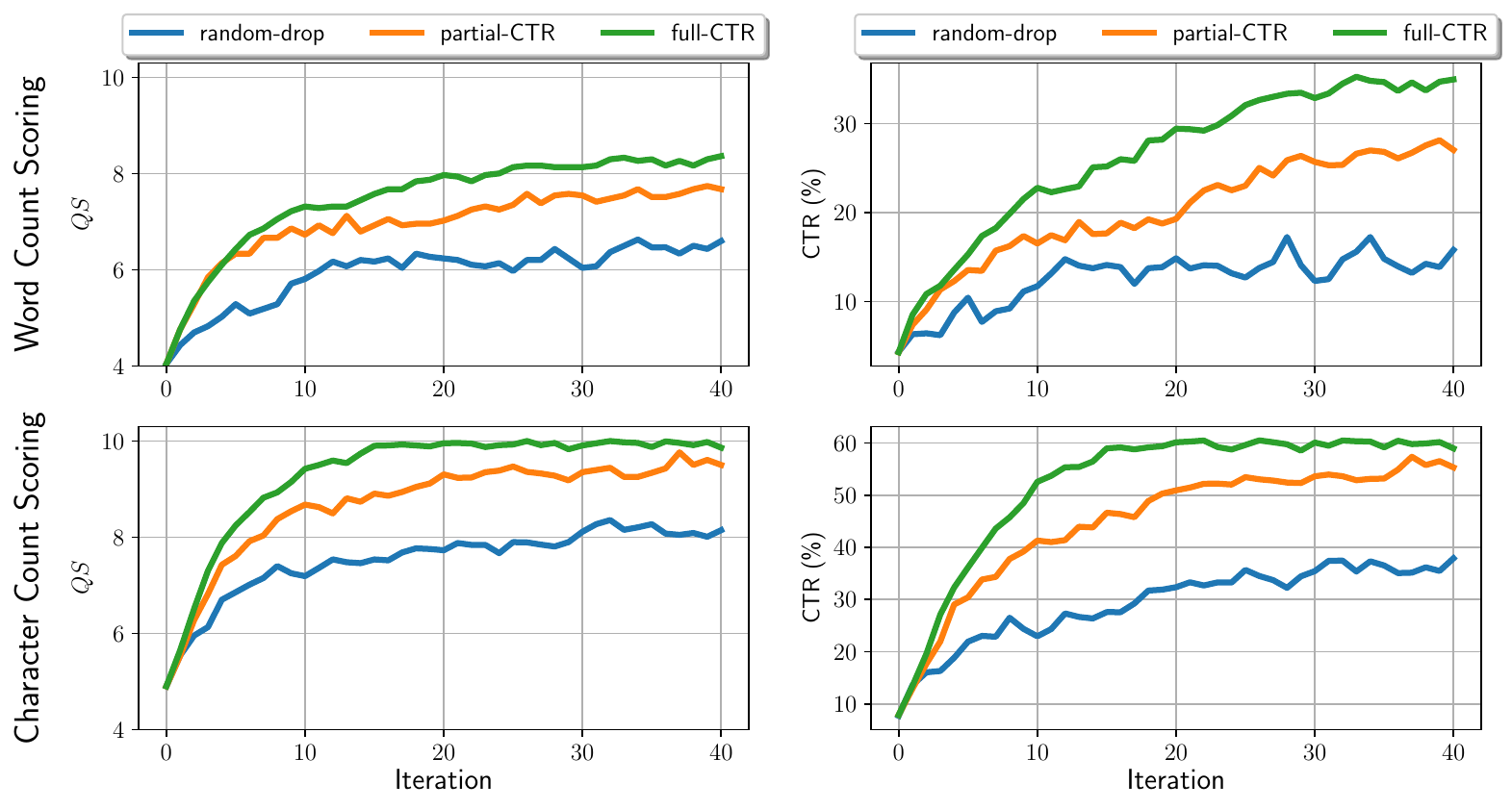}
    \caption{Average questions scores and CTRs for the \randomdrop, \partialctr, and \ctr methods in the artificial setup with length based scoring.}
    \label{fig:length_exp}
\end{figure*}

Instead of scoring the questions with LLMs based on their relevance to the personas as in \S\ref{subsec:relevance_scoring}, we also evaluate how different methods perform when each question is scored based on its length, measured in number of words and number of characters.
Word and character based scoring of a question $Q$ is calculated as following:
\[
\begin{aligned}
    \textsc{qs-w}(q) &=  max(min((|q|_w - 4) \times \frac{9}{11} + 1, 10), 1) \\
    \textsc{qs-c}(q) &= max(min((|q|_c - 20) \times \frac{9}{55} + 1, 10), 1)
\end{aligned}
\]
\noindent where $|q|_w$ and $|q|_c$ are the number of words and characters in $q$, respectively.
These formulas map the word and character counts of a question to a score between 1 and 10 with a maximum length of 15 words and 75 characters and a minimum length of 4 words and 20 characters. 
This deterministic scoring ensures a clear and observable trend in the CTR values of the questions and allows us to test whether and LLM can learn this trend and optimize for it. 
This setup requires the LLMs to learn to ignore the semantics and focus on the length of the questions. 
Although one can use any type of input for this setup, for simplicity, we opt for using the same inputs (i.e., product categories) that we use in our main experiments for the shopping domain (\Cref{subsubsec:shopping_domain}).

Next, we investigate \randomdrop, \partialctr and \ctr methods on the artificial setup with scoring based on question length. 
Figure \ref{fig:length_exp} displays the average question scores and CTRs for the three methods across 40 iterations for word count and character count based scoring. We increased the number of iterations to 40 for this experiment to make sure that all methods converged. 
The trend is similar and consistent for both scoring methods.
In early iterations the average question length goes up for all methods including the \randomdrop baseline, which is suprising since the \randomdrop baseline does not use the CTR information to drop shorter questions at each iteration.
We argue that this increase in the average question length is due to the fact that the prompt that is used to generate the initial 5 questions is different from the prompt that is used to generate the new questions at every iteration and, although it is not stated in the prompt, the iterative prompt, for some unclear reason, leads to longer questions.
The improvement for the \partialctr is expected since it uses the CTR data to drop shorter questions at every iteration while keeping the longer questions and the performance increases whenever a longer question is generated.
We also observe from the \ctr method that providing the CTR data to the LLM leads to a faster and greater increase in the average question length throughout the iterations.
This indicates that the GPT-4 implicitly recognizes that the customers preferred longer questions over the shorter ones and it is more likely to generate longer questions at the next iterations than the \partialctr method.

\section{Persona-level Results} \label{sec:AllResults}

\subsection{E-Commerce}\label{appendix:shop}

\begin{figure*}[!htb]
    \centering
    \includegraphics[width=\textwidth]{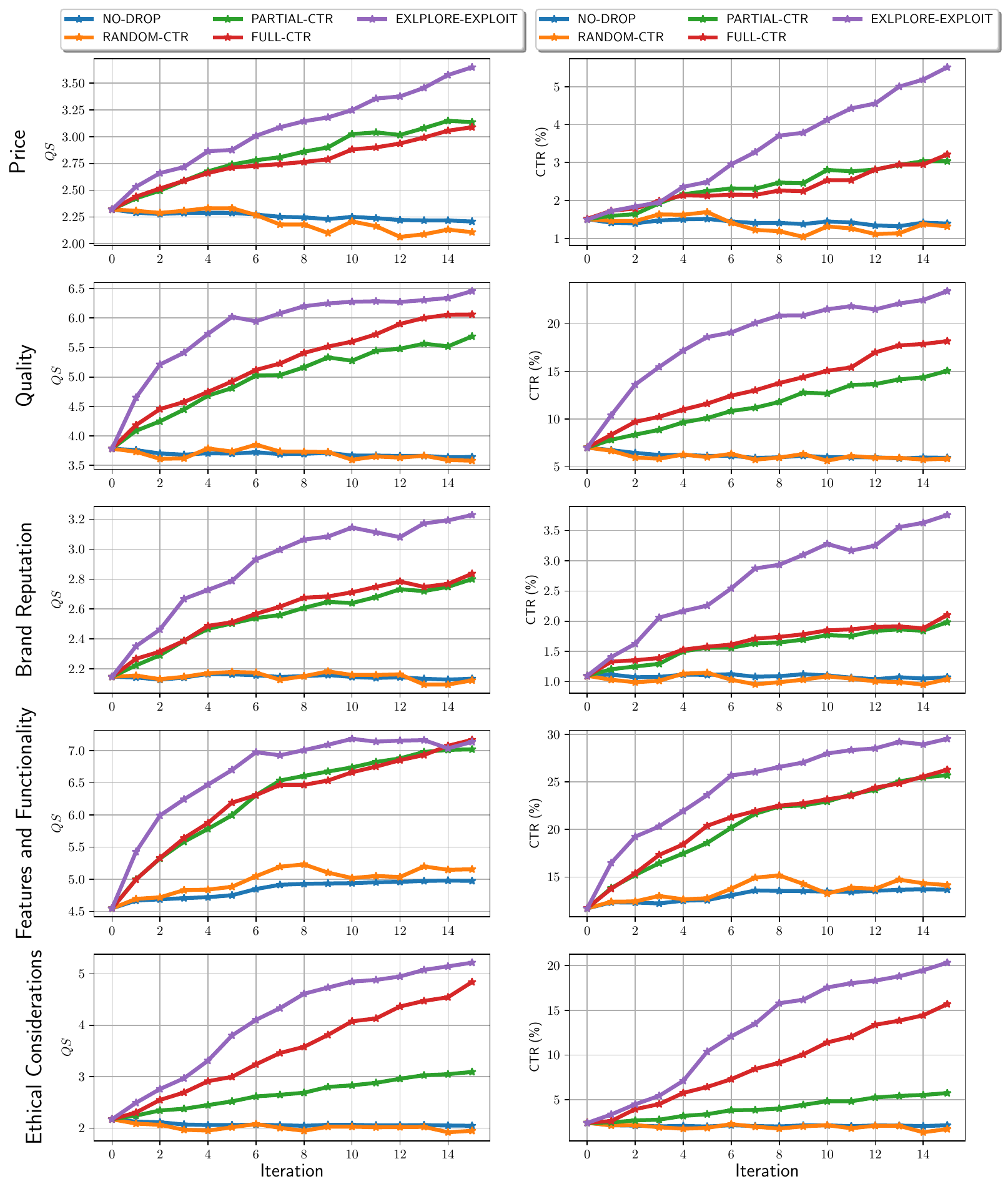}
    \caption{Average question scores and \ctr in the e-commerce domain for populations with single personas.}
    \label{fig:shopping_single_preference}
\end{figure*}

Figure \ref{fig:shopping_single_preference} shows the average question scores and CTRs of different methods in the e-commerce domain for populations with single personas. Consistent with the findings discussed in Section \ref{subsec:ctr_optimization_evaluation}, \ee demonstrates the most significant relative improvement in both question scores (between +1 and +3) and CTRs (between +2.5\% and +18\%). 
For three out of the five personas (\textit{Quality}, \textit{Features and Functionality}  and \textit{Ethical Considerations} ), \fullctr method performs comparable to the \ee method, indicating that even without an explicit explore-exploit instruction, LLM is able to find a good balance between exploring new topics and exploiting the best performing questions.
Note that the variance in achieved peak question scores and CTRs can be substantial across different personas. 
For all methods, achieved question scores and CTRs are substantially lower for the personas with \textit{Price} and \textit{Brand Reputation} preferences compared to the other three personas.
This is most likely because \textit{Price} and \textit{Brand Reputation} are more specific preferences compared to the other three, hence often they are not deeply explored in the question generation phase.
Overall, these results indicate that our proposed methods enhance the question generation process, consequently improving CTRs, for a diverse set of personas.

\subsection{General Knowledge}
\label{appendix:gen}

\begin{figure*}[!htb]
    \centering
    \includegraphics[width=\textwidth]{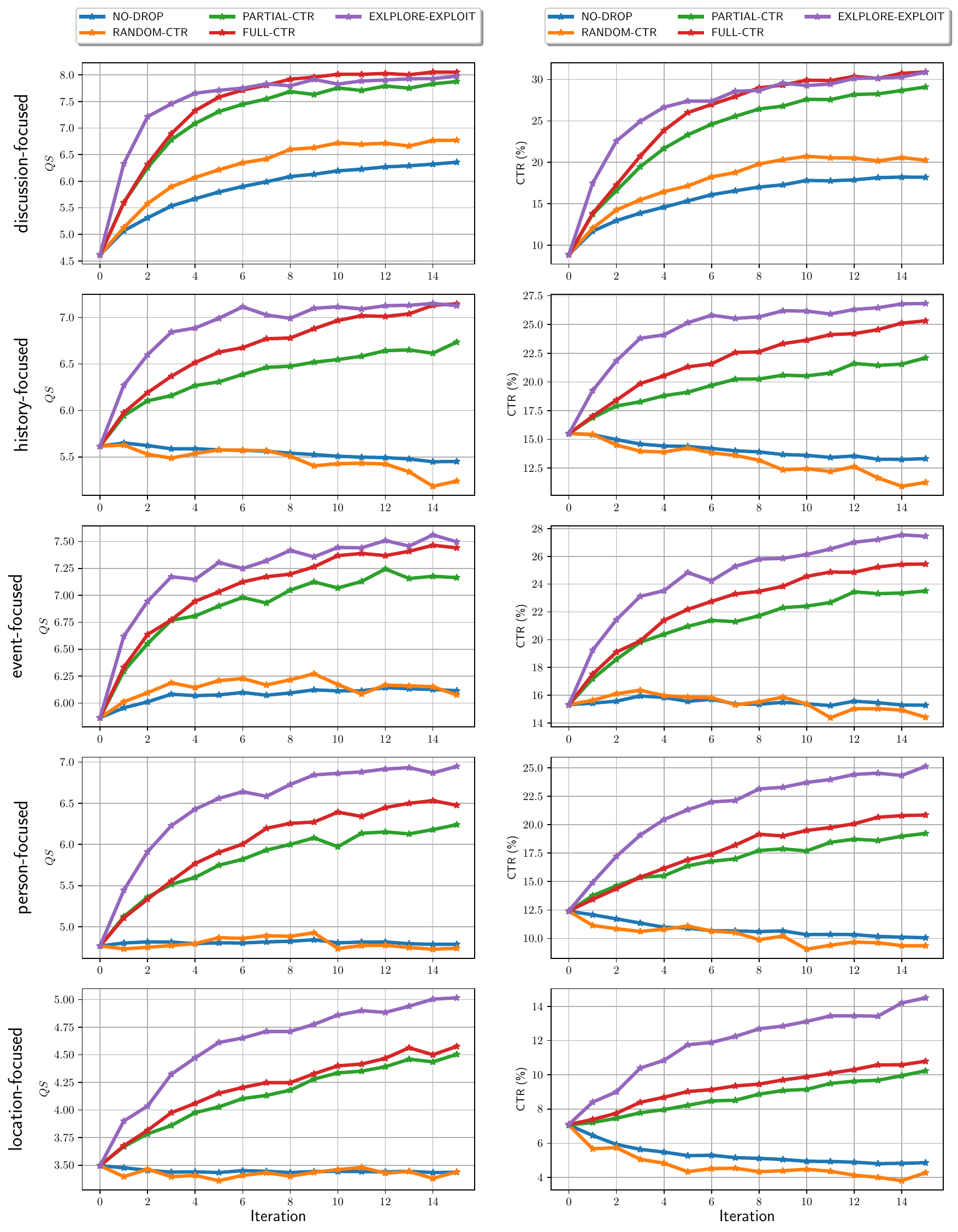}
    \caption{Average question scores and CTRs for the \partialctr, \ctr and \ee methods on general knowledge domain for personas with a single preference.}
    \label{fig:wiki_single_focus_all}
\end{figure*}

Figure~\ref{fig:wiki_single_focus_all} illustrates the average question scores and CTRs  in the general knowledge domain, focusing on five personas with a single preference. Similar to the observations in \S\ref{appendix:shop},  \ee demonstrates the most notable improvement in question scores (between +1 and +4) and CTRs (between +3 and +25) when comparing the last iteration to the initial one. 
Similar to the personas in the E-Commerce domain, we observe significant variations in the ultimate question scores and CTRs across different personas. 
For instance, by iteration 10, the ``event-focused'' persona could attain approximately a 30\% CTR, whereas the maximum CTR achieved for the ``location-focused'' persona after all iterations was only 6\%. 
This is likely because for many of the Wikipedia articles (e.g., Tabata Training) location related questions are not typical and hence LLM does deeply not explore generating questions that are relevant to this persona.
These results demonstrate that our proposed methods effectiveness could generalize to different domains.

\subsection{Average CTR Values} Table \ref{tab:avg_ctr} presents the average and last CTR values after 15 iterations for all of the tested methods and populations in this study.  In terms of average CTRs, \ee outperforms all other methods in 12 out of 13 cases, with \fullctr winning once.  For ``LOCATION-FOCUSED'' persona, \fullctr achieves 0.2\% higher CTR than \ee. 
Considering the CTRs at the last iteration, \ee outperforms in 11 out of 13 cases, with \fullctr winning twice. Notably, with the ``Event-Focused'' persona, \ee surpasses \fullctr by approximately 8.8\% at the last iteration. 
Overall, the persona-level results are consistent with our observations in \S\ref{appendix:shop} and \S\ref{appendix:gen}.

\begin{table*}[!htb]
    \centering
    \resizebox{.9\textwidth}{!}{
    \begin{tabular}{l rrrrrrrrrr}
    \toprule
    \multirow{2}{*}{Personas} & \multicolumn{2}{c}{\nodrop} & \multicolumn{2}{c}{\randomdrop}  & \multicolumn{2}{c}{\partialctr} & \multicolumn{2}{c}{\fullctr} & \multicolumn{2}{c}{\ee} \\
     & Avg. & Last & Avg. & Last & Avg. & Last & Avg. & Last & Avg. & Last \\
    \midrule
    \textsc{Price} & 1.4\% & 1.4\% & 1.4\% & 1.3\% & 2.4\% & 3.0\% & 2.3\% & 3.2\% & 3.4\% & 5.5\% \\
    \textsc{Quality} & 6.2\% & 6.0\% & 6.1\% & 5.9\% & 11.4\% & 15.1\% & 13.3\% & 18.2\% & 18.5\% & 23.4\% \\
    \textsc{Brand Reputation} & 1.1\% & 1.1\% & 1.0\% & 1.0\% & 1.6\% & 2.0\% & 1.7\% & 2.1\% & 2.7\% & 3.8\% \\
    \textsc{Features \& Functionality} & 13.1\% & 13.7\% & 13.6\% & 14.1\% & 20.4\% & 25.7\% & 20.8\% & 26.3\% & 24.4\% & 29.5\% \\
    \textsc{Ethical Considerations} & 2.1\% & 2.2\% & 2.0\% & 1.7\% & 4.1\% & 5.8\% & 8.9\% & 15.7\% & 12.7\% & 20.3\% \\
    \midrule
    \textsc{3 Preference} & 2.9\% & 2.8\% & 2.9\% & 2.7\% & 4.9\% & 6.2\% & 5.8\% & 7.3\% & 7.5\% & 9.5\% \\
    \midrule
    \midrule
    \textsc{Discussion-Focused} & 15.8\% & 18.2\% & 17.8\% & 20.3\% & 23.5\% & 29.1\% & 25.4\% & 30.9\% & 26.4\% & 30.9\% \\
    \textsc{History-Focused} & 14.1\% & 13.3\% & 13.2\% & 11.3\% & 19.7\% & 22.1\% & 21.9\% & 25.3\% & 24.5\% & 26.8\% \\
    \textsc{Event-Focused} & 15.5\% & 15.3\% & 15.4\% & 14.4\% & 21.1\% & 23.5\% & 22.4\% & 25.4\% & 24.4\% & 27.4\% \\
    \textsc{Person-Focused} & 10.8\% & 10.1\% & 10.3\% & 9.4\% & 16.8\% & 19.2\% & 17.8\% & 20.8\% & 21.4\% & 25.1\% \\
    \textsc{Location-Focused} & 5.4\% & 4.9\% & 4.7\% & 4.3\% & 8.7\% & 10.2\% & 9.3\% & 10.8\% & 11.8\% & 14.5\% \\
    \bottomrule
    \end{tabular}}
    \caption{\ctravg and \ctr on \QPI{15} for the \partialctr, \fullctr, \ee, \randomdrop, and \nodrop methods for different populations.}
    \label{tab:avg_ctr}
\end{table*}

\clearpage

\section{Generated Questions}
\label{sec:examples}

In Tables~\ref{tab:questions_shopping} and \ref{tab:questions_wiki}, we show the questions that are from the 1st iteration and last iteration for the \ee method for E-commerce and Wikipedia domains, respectively. We could clearly see that after refinement, the generated questions are much more relevant to the Persona and Topic. For example, when Persona is ``Quality'' and Topic is ``Cookware Sets'', irrelevant questions like ``How many pieces are typically included in a cookware set?'' are replaced with more relevant questions at the last iteration.

\begin{table*}[htbp]
    \centering
    \resizebox{\columnwidth}{!}{\begin{tabular}{c c p{0.5\columnwidth} p{0.5\columnwidth}}
    \toprule
    Persona & Topic & Initial Questions & After \ee \\
    \midrule
    \multirow{5}{*}{\rotatebox[origin=c]{90}{Quality}}  & \multirow{5}{*}{\rotatebox[origin=c]{90}{Cookware Sets}} & What materials are commonly used in cookware sets?  & What materials are commonly used in cookware sets? \\
    & & How many pieces are typically included in a cookware set? & What factors affect the durability of different cookware set materials? \\
    & & Are non-stick cookware sets safe for health? & What is the optimal thickness for stainless steel cookware for even heat distribution? \\
    & & Can cookware sets be used on induction cooktops? & Are copper cookware sets prone to tarnishing over time? \\
    & & How do I properly care for a stainless steel cookware set? & How do different cookware set materials resist wear and tear over time? \\
    \midrule
    \multirow{5}{*}{\rotatebox[origin=c]{90}{Features and Functionality}}  & \multirow{5}{*}{\rotatebox[origin=c]{90}{Lighting \& Ceiling Fans}} & What types of lighting fixtures are available for home use?  & How can smart lighting systems be optimized for remote control and automation? \\
    & & How do ceiling fans improve air circulation? & What are the steps to integrate smart lighting with home voice assistants? \\
    & & What are the differences between LED and incandescent bulbs? & What are the key features to look for in remote-controlled smart lighting systems? \\
    & & Can lighting be used to make a room appear larger? & What guidelines exist for the disposal of old or broken light bulbs? \\
    & & What is the average lifespan of a ceiling fan? & How do you program smart lighting for different time zones in a household? \\
    \midrule
    \multirow{5}{*}{\rotatebox[origin=c]{90}{Ethical Considerations}}  & \multirow{5}{*}{\rotatebox[origin=c]{90}{Spray Bottles}} & What are spray bottles typically used for? & How do I identify the recycling code on a spray bottle? \\
    & & How do spray bottles work? & How do I decode the recycling symbols on my spray bottle? \\
    & & What materials are spray bottles made from? & What are the steps for disassembling a spray bottle before recycling? \\
    & & Are spray bottles recyclable? & Are there eco-friendly biodegradable options for spray bottles? \\
    & & Can spray bottles be reused? & What items should be removed before recycling a spray bottle? \\
    \bottomrule
    \end{tabular}}     
    \caption{Questions in the initial item pool and after 15 iterations with \ee method for some personas and topics from the E-Commerce domain.}
    \label{tab:questions_shopping}
\end{table*}

\begin{table*}[htbp]
    \centering
    \resizebox{\columnwidth}{!}{\begin{tabular}{c c p{0.5\columnwidth} p{0.5\columnwidth}}
    \toprule
    Persona & Topic & Initial Questions & After \ee \\
    \midrule
    \multirow{5}{*}{\rotatebox[origin=c]{90}{Event-Focused}}  & \multirow{5}{*}{\rotatebox[origin=c]{90}{Petra}} & What civilization built Petra and when was it established?  & How did Petra's rediscovery to the Western world occur? \\
    & & How and when did Petra become a UNESCO World Heritage Site? & What prompted Johann Ludwig Burckhardt to seek out and identify Petra in 1812? \\
    & & For what purpose was Petra primarily used? & What led to the systematic exploration of Petra in the 19th century?  \\
    & & What are some of the most notable architectural features of Petra? & How did religious practices shape Petra's architectural landscape? \\
    & & Where is Petra located? & What specific events marked Petra's introduction to the global scholarly community? \\
    \midrule
    \multirow{5}{*}{\rotatebox[origin=c]{90}{Person-Focused}}  & \multirow{5}{*}{\rotatebox[origin=c]{90}{Lake Baikal}} & Why is Lake Baikal considered unique in terms of biodiversity?  &How do contemporary Baikal indigenous practices reflect their spiritual connection to the lake? \\
    & & Are there any notable species endemic to Lake Baikal? & How does Lake Baikal feature in the oral histories of local indigenous groups? \\
    & & Where is Lake Baikal located? & How have indigenous traditions shaped the conservation of Lake Baikal? \\
    & & How deep is Lake Baikal? & How have indigenous narratives influenced Lake Baikal's environmental policies and protections? \\
    & & What is the age of Lake Baikal? & What underwater features characterize Lake Baikal's unique topography? \\
    \midrule
    \multirow{5}{*}{\rotatebox[origin=c]{90}{Location-Focused}}  & \multirow{5}{*}{\rotatebox[origin=c]{90}{Kabuki}} & When and where did Kabuki originate? & What regions of Japan were instrumental in the development of Kabuki theater? \\
    & & What is the significance of makeup in Kabuki? & How did different regions in Japan contribute to Kabuki's theatrical traditions? \\
    & & How are roles distributed in Kabuki theatre? & When and where did Kabuki originate? \\
    & & What is Kabuki? & How have regional variations influenced the evolution of Kabuki's performance style? \\
    & & What are the key features of a Kabuki performance? & What traditional instruments are used in Kabuki music accompaniment? \\
    \bottomrule
    \end{tabular}}     
    \caption{Questions in the initial item pool and after 15 iterations with \ee method for some personas and topics from the Wikipedia domain.}
    \label{tab:questions_wiki}
\end{table*}

\end{document}